\pdfoutput=1
\documentclass[11pt,letterpaper]{article}
\usepackage[charter]{altum_arxiv}   %
\usepackage{cuted} %

\usepackage{currfile} %

\usepackage{caption} %

\usepackage{xspace}
\usepackage{pifont}   %
\definecolor{okgreen}{RGB}{0,140,60}\definecolor{badred}{RGB}{205,40,40}   %

\setlist{nosep, leftmargin=*}

\newcommand{\twoline}[1]{\begin{tabular}[b]{@{}c@{}}#1\end{tabular}}

\title{FrameSeg3D: Learned Local Frames for Segmenting\\Small Parts on High-Resolution 3D Scans}
\newcommand{\AuthorNames}{%
 Xinwen Zhang\textsuperscript{1,\dag}\quad
 Tongyan Wang\textsuperscript{2}\quad
 Huawei Lin\textsuperscript{3,\corrmark}}
\newcommand{\AuthorAffils}{%
 \textsuperscript{1}Altum Technology\quad
 \textsuperscript{2}Purdue University\quad
 \textsuperscript{3}Rochester Institute of Technology}
\newcommand{\AuthorNotes}{\textsuperscript{\dag}Project Lead\qquad\textsuperscript{\corrmark}Corresponding Author: \href{mailto:hl3352@rit.edu}{hl3352@rit.edu}}
\author{\AuthorNames\\\AuthorAffils}

\begin{document}
\maketitle
\begin{strip}
\centering
\includegraphics[width=\linewidth]{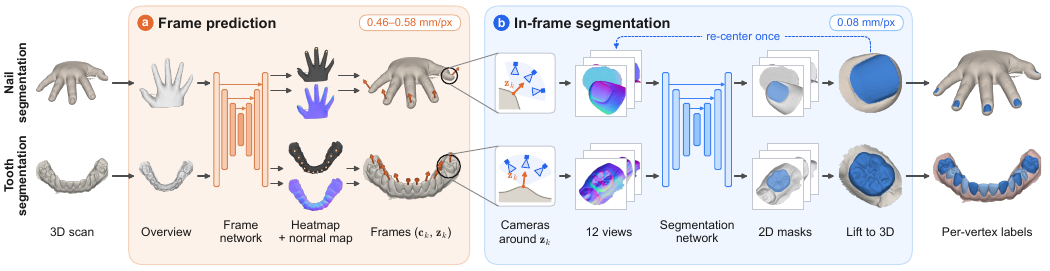}
\captionsetup{hypcap=false}
\captionof{figure}{\textbf{Overview of FrameSeg3D.} \textbf{(a) Frame prediction.} A frame network sees a low-resolution,
top-down overview of the whole scan. It predicts a heatmap peaking at each part's center $\mathbf{c}_k$, a normal map that
gives the part's direction $\mathbf{z}_k$ there (color), and a name map (not shown). Together they
give each part's local frame (arrows). The circled thumb's nail is tilted $49^\circ$ from the mat. \textbf{(b) In-frame
segmentation.} Twelve cameras within $60^\circ$ of $\mathbf{z}_k$ render each part at 0.08\,mm per pixel, so it is seen
centered, facing the camera, and at a fixed scale. A segmentation network segments every view, the masks are lifted back
to the vertices, and the frame is re-centered once. Each task trains its own frame network; one segmentation network
serves both. All panels but the camera insets are real outputs of FrameSeg3D on test scans.}
\captionsetup{hypcap=true}
\label{fig:teaser}%
\end{strip}

\begin{abstract}
Segmenting small parts on high-resolution 3D scans, such as nails on hand scans and teeth on intraoral scans, poses
three challenges:
1) each part covers a tiny fraction of the scan, 2) its boundary is only a subtle change of height and
curvature, and 3) it faces its own direction. Existing methods fall into two groups, and both fall short. The first
group processes the whole scan at once and sees each part too coarsely. Point and mesh networks must shrink the scan to
fit in memory, so a small part keeps only a few of its points. Multi-view methods render images of the whole object, so
each part fills only a few pixels and a tilted part is seen from the side. The second group first finds each part and
then segments it up close. But these methods find the parts with knowledge about one kind of part, for example the
shape of the dental arch for teeth. Such knowledge does not carry over to other kinds of parts. We find that accuracy is limited by how a part is seen, from which
direction and at what scale, not by how many points land on it. We therefore present
FrameSeg3D. It first predicts a local frame for every part, its center and the direction it faces, from a low-resolution
view of the whole scan. It then segments the part in views around its frame normal at the scanner's resolution, so
covering the whole scan no longer costs resolution. Because every part arrives centered, facing the camera, and at a
fixed scale, one segmentation network can serve different kinds of parts and learn from few labeled masks. We test
FrameSeg3D on two kinds of parts, nails and teeth. It outperforms state-of-the-art methods on hand scans (mean Dice 0.970 on nails) and on the Teeth3DS+ test fold
(per-tooth IoU 96.5), with the most accurate tooth boundaries (boundary IoU 70.5).
\end{abstract}
\section{Introduction}
\label{sec:intro}

\textbf{Small parts on high-resolution scans.} 3D scanners capture surfaces in fine detail, from large
statues~\cite{levoy2000michelangelo} to the inside of the mouth~\cite{mangano2017intraoral}, and many scans are taken to
measure one small part of a larger surface. Examples are a fingernail on a hand scanned for a custom-fit artificial
nail~\cite{gedamu2022nails,deng2026nail} and a tooth on an intraoral scan used to plan a
crown~\cite{miyazaki2009cadcam,benhamadou2022teeth3ds,benhamadou2023teethseg}. Parts in shape
collections~\cite{chen2009benchmark,chang2015shapenet,yi2016shapenetpart,mo2019partnet} are large, have sharp edges, and
often come on aligned shapes. In contrast, these parts are tiny: a nail holds about 6,000 of a hand
scan's two million vertices (0.3\%). Their boundaries are subtle changes of height and curvature, with no color to
help. Each part also faces its own direction. On a hand lying flat, the thumb's nail faces a median $56^\circ$ away from
the mat. Segmenting them vertex by vertex requires seeing each part at the scanner's resolution and from roughly the
direction it faces.

\textbf{Existing methods.} No existing family of methods sees every part at the scanner's resolution and from the
direction it faces. \emph{Point and mesh
networks}~\cite{qi2017pointnet,qi2017pointnetpp,wang2019dgcnn,zhao2021pointtransformer,wu2024ptv3,hanocka2019meshcnn,sharp2022diffusionnet,lian2020meshsegnet}
fit the whole scan into a fixed budget of samples. On a hand, this leaves 41 to 671 samples on a nail, too few to
trace its boundary. \emph{Multi-view
methods}~\cite{su2015mvcnn,kundu2020virtual,benhamadou2023teethseg,lu2026teethsam,xi2025crosstooth}, including zero-shot
ones built on 2D foundation models~\cite{liu2023partslip,abdelreheem2023satr,tang2024samesh,zhuang2026tsegagent}, place
their cameras for the whole object. A part gets few pixels, and a tilted part is seen at a grazing angle. On our hand scans, finer
pixels even make a whole-hand render slightly worse. \emph{Two-stage dental methods} first find each tooth, then
segment it in a 3D crop~\cite{cui2021tsegnet,benhamadou2023teethseg,vannistelrooij2025toothinstancenet}, a
landmark-aligned patch~\cite{lee2025render2seg}, a 2D parameterization~\cite{rekik2025tseglab}, or views placed by the
dental arch or the occlusal plane~\cite{huang2024iossam,kubik2024lmvsegrnn}. They come closest to seeing each part on
its own, but they have two limits. They find the teeth through the structure of the dentition, such as the arch,
landmarks and tooth positions, which other parts lack. And their crops or views are set by the whole jaw, not by the
direction each part faces. On our nails, ToothGroupNetwork, the winner of the 3DTeethSeg'22
challenge~\cite{benhamadou2023teethseg}, reaches a mean Dice of only 0.797.

\textbf{How a part is seen.} We argue that small parts are limited by how they are seen, from which direction and at
what scale, rather than by how many samples land on them. With about as many pixels on a nail, views around the nail's
own normal reach a mean Dice of 0.970, against 0.956 for views of the whole hand. In images and medical volumes, small
objects are handled by zooming in~\cite{xia2016zoom,gao2018dynamic,najibi2019autofocus,zhou2017fixedpoint} or by
processing each object at a fixed scale~\cite{he2017mask,singh2018snip,singh2018sniper}. On a 3D surface, a small part must also be turned toward the camera. Views centered
on a nail but placed around the mat's normal fall to 0.927 on nails tilted more than $45^\circ$, against 0.965 around
the nail's own normal. So each part needs a \emph{frame}: its center and the direction it faces. Finding the frame is hard. It is defined by the part
itself, yet needed before the part is segmented. So local reference frames computed from the surface around a
point~\cite{tombari2010shot,tatarchenko2018tangent,huang2019texturenet} cannot provide it, and learned
alignment~\cite{jaderberg2015spatial,qi2017pointnet,sajnani2022condor} and view selection~\cite{hamdi2021mvtn} act on the
whole input, not on each part. It must be found in a view where the part is small. And it must not
rely on the structure of one kind of part, such as the dental arch.

\textbf{Segmenting each part in its own frame.} We present FrameSeg3D, which predicts a coarse frame for every part and
segments each part in its own frame (\cref{fig:teaser}). \emph{Frame prediction} looks at the whole scan once, at low
resolution. A frame network~\cite{ronneberger2015unet} on a top-down render finds every part as a heatmap
peak~\cite{zhou2019objects}, names it, and predicts the direction it faces.
Unlike two-stage dental methods, it needs no arch curve, landmark detector, or shape model, and no extra labels: we
compute the training frames from the segmentation masks. \emph{In-frame segmentation} renders each part from 12
directions around its own frame normal at 0.08\,mm per pixel, finer than the scan's vertex spacing. So the pixel size
no longer depends on the size of the scan, and the view direction no longer depends on where the part lies. A 2D
network segments every view, and we lift the masks back to the vertices~\cite{kundu2020virtual}. A coarse frame is enough, because the segmentation network is trained on perturbed frames
and we re-center the frame once on the first result. Every part then arrives centered, facing the camera, and at a
fixed scale. So one segmentation network serves nails and teeth alike and learns from few labels. FrameSeg3D reaches
a mean Dice of 0.970 on the nails of held-out people, ahead of point, mesh, and multi-view networks. On the Teeth3DS+
test fold~\cite{benhamadou2022teeth3ds}, it has the best per-tooth IoU (96.5) and the most accurate tooth boundaries.
With both stages trained on a quarter of the hands or a tenth of the patients, it still beats the strongest baselines
trained on all of them.

\textbf{Contributions.} Our contributions are as follows:
\begin{itemize}
  \item We propose FrameSeg3D, which segments small parts on high-resolution 3D scans in learned local frames. It
        predicts a coarse frame for each part, its center and the direction it faces, from a low-resolution view of the
        whole scan, with no domain prior and no labels beyond the segmentation masks. It then segments 2D views around
        the frame normal at the scanner's resolution and lifts the masks back to the vertices.
  \item We observe, in controlled experiments, that small-part accuracy is limited by viewing geometry, not by the number
        of samples. At matched pixels per part and the same crop center, the part's own frame wins. On hand scans,
        finer pixels help only inside the frame, and the gain grows with the part's tilt.
  \item We present a segmentation network that does not depend on the part and learns from few labels. Trained on
        nails and teeth together, it matches two dedicated networks.
  \item Our experimental results show that FrameSeg3D is the most accurate method on nails and on the Teeth3DS+ test
        fold, and draws the most accurate tooth boundaries on both Teeth3DS+ test sets. On the 3DTeethSeg'22 challenge
        test set, it trails ToothInstanceNet~\cite{vannistelrooij2025toothinstancenet} by 0.4 per-tooth IoU, mostly on
        third molars, which are rare in the training fold, so our frame network often places no frame on them.
\end{itemize}

\section{Related Work}
\label{sec:related}

\textbf{Segmenting the whole scan.}
Point networks~\cite{qi2017pointnet,qi2017pointnetpp,wang2019dgcnn,zhao2021pointtransformer,wu2024ptv3,wu2025sonata,yue2026litept}
and mesh networks~\cite{hanocka2019meshcnn,sharp2022diffusionnet} segment a scan from sampled points, faces or vertices.
On intraoral scans (Teeth3DS+ and the 3DTeethSeg'22 challenge~\cite{benhamadou2022teeth3ds,benhamadou2023teethseg}),
MeshSegNet~\cite{lian2020meshsegnet} and TSGCNet~\cite{zhang2021tsgcnet} label a downsampled mesh, and recent methods add boundary-preserving
sampling~\cite{xi2025crosstooth} or a B-spline model of the dental arch~\cite{wei2026bspline}. All of them fit a scan
into a fixed sample budget, so a small part gets samples only in proportion to its share of the surface. Multi-view
methods render the shape, segment the images, and fuse the 2D predictions on the
surface~\cite{su2015mvcnn,kalogerakis2017projective,kundu2020virtual}, also with 2D foundation models for zero-shot part
segmentation~\cite{liu2023partslip,abdelreheem2023satr,yang2024sampart3d,tang2024samesh,liu2025partfield}. For teeth,
they render the jaw from random~\cite{benhamadou2023teethseg} or fixed
viewpoints~\cite{lu2026teethsam,zhuang2026tsegagent,xi2025crosstooth}. Their cameras are placed for the whole object, so a
part's pixel count and viewing angle depend on where it lies. FrameSeg3D keeps the render-and-lift pipeline but places
the cameras per part, around its own normal, at a fixed metric resolution.

\textbf{Finding the parts first.}
Zooming into a region is an established idea for images~\cite{xia2016zoom,he2017mask}. Two-stage dental methods bring
it to intraoral scans. They locate the teeth and segment each one in a 3D
crop~\cite{cui2021tsegnet,qiu2022darch,benhamadou2023teethseg,vannistelrooij2025toothinstancenet}, a landmark-aligned
patch~\cite{lee2025render2seg}, a 2D parameterization of the crown~\cite{rekik2025tseglab}, or views along the dental
arch~\cite{huang2024iossam}. Closest to our design, LMVSegRNN segments each tooth from views that circle the occlusal
axis above two detected landmarks~\cite{kubik2024lmvsegrnn}. These methods rely on dental structure to find the teeth,
and their crops or views are set by the whole jaw, not by the direction each tooth faces. For nails, prior work made
custom-fit artificial nails from 3D scans~\cite{gedamu2022nails} and grew the nail region on a finger
scan~\cite{deng2026nail}, our classical baseline. Canonical frames are common in learning: alignment
modules align a whole input~\cite{jaderberg2015spatial,qi2017pointnet}, local reference frames make descriptors
invariant to rotation~\cite{tombari2010shot}, tangent convolutions view each point along its
normal~\cite{tatarchenko2018tangent}, Frustum PointNets rotate each detected object~\cite{qi2018frustum}, and FrameNet
predicts a frame at every pixel~\cite{huang2019framenet}. FrameSeg3D instead learns one frame per part, its center and
facing direction, from a coarse view of the whole scan, with no dental structure. The frame places the
cameras, the scanner's resolution sets the pixel size, and the frame labels come for free from the segmentation masks.
\section{Method}
\label{sec:method}

\textbf{Problem setting.}
The input is a triangle mesh $\mathcal{M}=(\mathcal{V},\mathcal{F})$ from a 3D scanner, with vertices $\mathcal{V}$,
triangular faces $\mathcal{F}$, and no color. Vertex $i$ has
position $\mathbf{v}_i\in\mathbb{R}^3$, unit normal $\mathbf{n}_i$, area $a_i$ (a third of its faces' area),
and mean curvature $\kappa_i$. The output is a label $y_i\in\{0,1,\dots,K\}$ per vertex: background, or one of $K$ named parts, such as the five nails of a hand ($K=5$)
or the up to 16 teeth of a jaw ($K=16$). Part $k$ covers the vertex set
$\mathcal{S}_k\subset\mathcal{V}$. Its \emph{frame} $\mathcal{T}_k=(\mathbf{c}_k,\mathbf{z}_k)$ is its area-weighted
centroid and mean normal, that is, where the part lies and the direction its surface faces on average:
\begin{equation}
  \mathbf{c}_k=\frac{\sum_{i\in\mathcal{S}_k}a_i\mathbf{v}_i}{\sum_{i\in\mathcal{S}_k}a_i},\qquad
  \mathbf{z}_k=\frac{\sum_{i\in\mathcal{S}_k}a_i\mathbf{n}_i}{\bigl\|\sum_{i\in\mathcal{S}_k}a_i\mathbf{n}_i\bigr\|}.
  \label{eq:frame}
\end{equation}
We compute ground-truth frames from the labels with \cref{eq:frame}. A rotation $\mathbf{R}_k=[\mathbf{x}_k\;
\mathbf{y}_k\;\mathbf{z}_k]^{\top}\in SO(3)$ adds two in-plane axes $\mathbf{x}_k,\mathbf{y}_k$, chosen freely. It maps a point to frame
coordinates, $\mathbf{v}^{(k)}=\mathbf{R}_k(\mathbf{v}-\mathbf{c}_k)$. \Cref{eq:frame} also shows the difficulty: the
frame depends on $\mathcal{S}_k$, the set we want to find.

\textbf{Pipeline.}
FrameSeg3D breaks this circle with two networks (\cref{fig:teaser}). Frame prediction $\Phi$ estimates every frame from a
low-resolution view of the whole scan. In-frame segmentation $\Psi$ then segments each part in its estimated frame, one part at a time:
\begin{equation}
  \{\hat{\mathcal{T}}_k\}_{k=1}^{K}=\Phi(\mathcal{M}),\qquad
  \mathbf{p}_k=\Psi\bigl(\mathcal{M},\hat{\mathcal{T}}_k\bigr)\in[0,1]^{|\mathcal{V}|},
  \label{eq:pipeline}
\end{equation}
where hats mark estimates and $p_{k,i}$ is the probability that vertex $i$ belongs to part $k$. \Cref{sec:s1,sec:s2} describe $\Phi$ and
$\Psi$, and \cref{sec:train} shows how $\Psi$ copes with $\Phi$'s errors.

\textbf{Why a frame.}
Consider an orthographic view along a direction $\mathbf{d}$ with pixel size $\delta$. A surface patch of area $A$ with
normal $\mathbf{n}$ is seen at the angle $\gamma=\angle(\mathbf{n},\mathbf{d})$. The patch gets $N$ pixels, spaced
$\Delta s$ apart on the surface along the direction in which it tilts away from the camera:
\begin{equation}
  N=\frac{A\cos\gamma}{\delta^{2}},\qquad \Delta s=\frac{\delta}{\cos\gamma}.
  \label{eq:sampling}
\end{equation}
In a view of the whole object, the object's size sets $\delta$, and where the part lies sets $\gamma$. A hand lying flat shows this
well: seen from above, the thumb's nail is at $\gamma\approx56^\circ$, so $\Delta s\approx1.8\,\delta$, and its far side is
compressed further. Views within a cone around $\mathbf{z}_k$ bound $\gamma$ wherever the part
lies, and we can set $\delta$ for the part alone, at the scanner's resolution. The frame says where to look and from
which direction: it sets both the crop and the cone of views.

\subsection{Frame prediction}
\label{sec:s1}

\textbf{Goal.}
Frame prediction $\Phi$ takes the whole scan and returns a name $k$ and a frame
$\hat{\mathcal{T}}_k=(\hat{\mathbf{c}}_k,\hat{\mathbf{z}}_k)$ for every part present. It must find every part,
including tilted ones, and name it correctly. The frame only needs to be close: in-frame segmentation tolerates a center
off by 4\,mm or a normal off by $20^\circ$ (\cref{sec:ablations}).

\textbf{Part size.}
One number, the part size $D$, sets every length in frame prediction and the crop radius in \cref{sec:s2}. We set $D$ to the median, over the labeled parts of the training scans, of twice the 90th percentile of the distance from a
part's centroid to its vertices ($D=11.5$\,mm for nails, 9.2\,mm for teeth). Twice the 90th percentile approximates a
part's diameter while ignoring stray vertices, and the median over parts ignores unusual ones. So $D$ is measured from
the labels, not tuned. Nothing else in the method depends
on the kind of part: nails and teeth share every other setting.

\textbf{Overview image.}
We take the scan's up direction $\mathbf{u}$ to be its area-weighted mean normal, because it needs no landmarks: for
a hand on a mat it is the mat's normal, and for a jaw it lies $9^\circ$ from the biting direction. We ray-cast the scan
orthographically along $\mathbf{u}$, center the image on the scan's centroid and align it with the scan's principal axes.
To keep every part about 20 pixels
across, enough to find it but not to segment it, we set the pixel size to $\delta_0=D/20$ (0.58\,mm for hands,
0.46\,mm for jaws). The overview image $\mathbf{I}^0$ has three channels: the height
above the local floor, the upward normal component $|\mathbf{n}\cdot\mathbf{u}|$, and coverage.

\textbf{Finding and naming parts.}
The frame network $f$ is a U-Net~\cite{ronneberger2015unet} with an ImageNet-pretrained~\cite{deng2009imagenet}
ResNet-34 encoder~\cite{he2016resnet}. A part may be missing, and neighbors can look alike, so $f$ does not
predict one heatmap per part. It predicts one center heatmap $\hat C$ for all parts, a map $\hat{\mathbf{S}}$ of the
$K$ part names, and a normal map $\hat{\mathbf{Z}}$. The heatmap finds the parts, so the name map is trained only on
the part pixels: small parts cover a tiny share of the image (a nail covers 0.1\% of a hand's image), and with
background pixels in its loss the network never learned to name them. We train the three outputs together with one loss,
a weighted sum of one standard term per output (a focal loss, a cross-entropy and a weighted L1 loss):
\begin{equation}
  \begin{aligned}
  \mathcal{L}_{\mathrm{frame}}&=\mathcal{L}_{\mathrm{focal}}(\hat C)+\mathcal{L}_{\mathrm{CE}}(\hat{\mathbf{S}})+0.1\,\mathcal{L}_{\mathbf{z}},\\
  \mathcal{L}_{\mathbf{z}}&=\frac{\sum_{\mathbf{q}}w(\mathbf{q})\,\bigl\|\hat{\mathbf{Z}}(\mathbf{q})-\mathbf{z}_{k(\mathbf{q})}\bigr\|_1}{\sum_{\mathbf{q}}w(\mathbf{q})},
  \end{aligned}
  \label{eq:frameloss}
\end{equation}
where $\mathbf{q}$ runs over the image's pixels, $\mathcal{L}_{\mathrm{focal}}$ is the heatmap loss of
CenterNet~\cite{zhou2019objects} (a focal loss that down-weights background pixels) on Gaussians at the projected
centroids, $\mathcal{L}_{\mathrm{CE}}$ is the cross-entropy, and $w(\mathbf{q})=1$ near the center of a part
$k(\mathbf{q})$, else 0. We follow CenterNet in scaling the Gaussians with the object size, here $D$, and in giving the
regression term the weight 0.1. The parts are the peaks of $\hat C$, and the Hungarian
algorithm~\cite{kuhn1955hungarian} gives them distinct names from $\hat{\mathbf{S}}$. \Cref{sec:app_impl} gives the remaining
constants, all set once as fractions of $D$ and shared by every kind of part.

\textbf{From a peak to a frame.}
A ray cast along $-\mathbf{u}$ through a peak hits the scan at the point $\mathbf{m}$. The frame center lies below
this point, by the typical depth of the part's centroid under the surface seen from above, that is, along $-\mathbf{u}$:
\begin{equation}
  \hat{\mathbf{c}}_k=\mathbf{m}-\bar d_k\,\mathbf{u},
  \label{eq:center}
\end{equation}
where $\bar d_k$ is the median depth of part $k$'s centroid below the surface seen from above, in the training scans
(close to zero for a thin part like a nail, inside the crown for a tooth). We read $\hat{\mathbf{z}}_k$ from $\hat{\mathbf{Z}}$ around
the peak. Before finding peaks, we average the outputs of three frame networks trained
with different random seeds, a small deep ensemble~\cite{lakshminarayanan2017ensembles}.

\subsection{In-frame segmentation}
\label{sec:s2}

\begin{figure}[t]
\centering
\includegraphics[width=\linewidth]{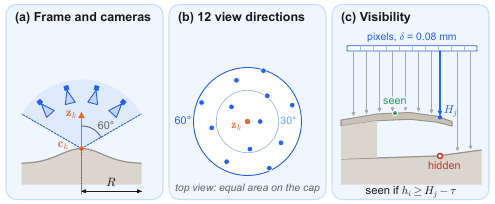}
\caption{\textbf{Geometry of in-frame segmentation.} (a) The crop is the surface within $R$ of the frame center
$\mathbf{c}_k$, seen by cameras within $60^\circ$ of the frame normal $\mathbf{z}_k$. (b) The 12 view directions
(\cref{eq:views}), seen from above. (c) A vertex takes a view's prediction only if the view sees it (\cref{eq:vis}); a
vertex under an overhang (here, a nail's free edge) is labeled by the other views, those that see it.}
\label{fig:geometry}
\end{figure}

\textbf{Goal.}
In-frame segmentation $\Psi$ takes the scan and one estimated frame $\hat{\mathcal{T}}_k$ and returns the probability
$p_{k,i}$ that each vertex $i$ near the frame belongs to part $k$. It must place the part's boundary to a fraction of a
millimeter, at the scanner's resolution. It sees only a crop around the frame, not the rest of the scan, so each part is handled on its own.

\textbf{Crop and views.}
The crop $\mathcal{C}_k$ is the connected piece of the scan within $R$ of $\hat{\mathbf{c}}_k$ that holds the topmost surface along $\hat{\mathbf{z}}_k$. We set $R=1.39D$, which is 16\,mm for nails and 12.8\,mm for teeth. This is the radius we first chose for nails, written
in units of $D$ so that one rule serves every kind of part. It leaves room for a frame center that is off by several
millimeters. We express it in frame coordinates
and view it from $J=12$ directions on a Fibonacci spiral~\cite{gonzalez2010fibonacci}, which cover the cap within $\theta_{\max}$ of
$\hat{\mathbf{z}}_k$ with equal area (\cref{fig:geometry}a,b). To sample a surface element facing $\hat{\mathbf{z}}_k$ at most $2\delta$ apart
(\cref{eq:sampling}) while still reaching the part's side walls, we set $\theta_{\max}=60^\circ$:
\begin{equation}
  \begin{aligned}
  \cos\theta_j&=1-(1-\cos\theta_{\max})\,\frac{j+\frac12}{J},\\
  \phi_j&=j\,\pi\bigl(3-\sqrt5\bigr),\qquad j=0,\dots,J-1.
  \end{aligned}
  \label{eq:views}
\end{equation}
View $j$ turns the crop by $\phi_j$ about $\hat{\mathbf{z}}_k$ ($\mathbf{R}_z$), tilts it by $\theta_j$ about the $x$
axis ($\mathbf{R}_x$), and looks down the new $z$ axis: vertex $i$ moves to
$\mathbf{w}^{(j)}_i=\mathbf{R}_x(\theta_j)\,\mathbf{R}_z(\phi_j)\,\mathbf{v}^{(k)}_i=(x_i,y_i,h_i)$, with height $h_i$. By
\cref{eq:sampling}, every view sees a surface element facing $\hat{\mathbf{z}}_k$ at $\gamma\le60^\circ$, however
the part is tilted.

\textbf{Rendering.}
Each view is an orthographic ray cast from above onto a $384\times384$ grid. To see the part at the scanner's resolution, we set its pixel
size to $\delta=0.08$\,mm, finer than the scan's vertex spacing. The field of view is then $W=384\,\delta=30.7$\,mm, wide
enough for the part and its surroundings. The ray through the pixel in row $r$ and column $s$ starts at
$\bigl((s+\frac12)\delta-\frac W2,\,(r+\frac12)\delta-\frac W2\bigr)$ and returns the height $H_j(r,s)$ of its first
hit. There, we interpolate the vertex attributes over the hit triangle into the six input channels of the network:
\begin{equation}
  \mathbf{I}_j(r,s)=\bigl[\,H_j/5,\ \mathbf{n}^{(j)},\ \kappa,\ \mathbf{1}_{\mathrm{hit}}\,\bigr]\in\mathbb{R}^{6},
  \label{eq:render}
\end{equation}
the height, the normal in view coordinates, the mean curvature, and coverage. Covering a whole scan at this pixel size would take far larger images: about $2500^2$ pixels per view for
a hand, some 42 times as many as our $384^2$ views.

\textbf{Segmentation and lifting.}
The segmentation network $g$ is a U-Net like $f$ with six input channels. It predicts a probability image
$\mathbf{P}_j=\sigma\bigl(g(\mathbf{I}_j)\bigr)$ for every view, where $\sigma$ is the sigmoid. A vertex projects to the continuous pixel position
$(\rho_i,\varsigma_i)=\bigl((y_i+\frac W2)/\delta-\frac12,\ (x_i+\frac W2)/\delta-\frac12\bigr)$ and is visible in view
$j$ if it lies on the rendered surface, that is, if no closer surface hides it from that view:
\begin{equation}
  \nu_{ij}=\mathbf{1}\bigl[\,h_i\ge H_j(\rho_i,\varsigma_i)-\tau\,\bigr],
  \label{eq:vis}
\end{equation}
where $\mathbf{1}[\cdot]$ is 1 if the condition holds, else 0, and $\tau$ is a tolerance. We set $\tau=0.25$\,mm, about one
vertex spacing; values from 0.1 to 0.5\,mm give the same accuracy (\cref{sec:app_nails}). This excludes vertices under an overhang or on
the far side of the part (\cref{fig:geometry}c). Each vertex averages
the bilinearly interpolated $\mathbf{P}_j$ over the views that see it, with equal weights,
\begin{equation}
  p_{k,i}=\frac{\sum_{j}\nu_{ij}\,P_j(\rho_i,\varsigma_i)}{\sum_{j}\nu_{ij}},
  \label{eq:lift}
\end{equation}
and $p_{k,i}=0$ if no view sees it.

\textbf{Assembly and re-centering.}
Part $k$ is the largest connected piece of $\{i\in\mathcal{C}_k: p_{k,i}>0.5\}$. Where the pieces of neighboring parts
overlap, a vertex goes to the part with the highest $p_{k,i}$. Each part then keeps its largest connected piece on the
scan. Finally, we re-estimate the frame from the segment with \cref{eq:frame} and segment the part once more. This
closes the loop of \cref{eq:frame}: a coarse frame gives a segment, and the segment gives the frame it implies.

\subsection{Training}
\label{sec:train}

\textbf{Goal.}
The two networks are trained separately, each with its own loss (\cref{eq:frameloss,eq:loss}). The segmentation network
must also handle errors in the predicted frames.

\textbf{Perturbed frames.}
We train it on ground-truth frames perturbed like frame-prediction errors:
\begin{equation}
  \tilde{\mathbf{c}}=\mathbf{c}+\epsilon_1\mathbf{x}+\epsilon_2\mathbf{y},\qquad
  \tilde{\mathbf{z}}=\mathbf{R}_{\mathbf{a}}(\alpha)\,\mathbf{z},
  \label{eq:perturb}
\end{equation}
where $\mathbf{x},\mathbf{y}$ are the frame's in-plane axes, $\epsilon_l\sim\mathcal{N}(0,\sigma_c^2)$ is clipped to
$\pm4$\,mm, $\sigma_c=1.5$\,mm, and
$\mathbf{R}_{\mathbf{a}}(\alpha)$ a rotation by $\alpha\sim\mathcal{U}(0^\circ,15^\circ)$ about a random axis
$\mathbf{a}\perp\mathbf{z}$. We set these ranges well above the errors we measure for frame prediction (median 0.4\,mm
and $4^\circ$; \cref{sec:ablations}), so the network also learns from poor frames. The in-plane axis is random too, so
the network relies on neither an exact frame nor its in-plane orientation. Each training crop is seen from one direction drawn uniformly over the same cap as the test views,
$\cos\theta\sim\mathcal{U}(\cos\theta_{\max},1)$ and $\phi\sim\mathcal{U}(0,2\pi)$.

\textbf{Loss.}
Let $\Omega$ be the pixels without an ignore label, with target $t$ and predicted probability $P$ at each. The loss
is binary cross-entropy plus soft Dice:
\begin{equation}
  \begin{aligned}
  \mathcal{L}_{\mathrm{seg}}={}&-\frac{1}{|\Omega|}\sum_{\Omega}\bigl[t\log P+(1-t)\log(1-P)\bigr]\\
  &+1-\frac{2\sum_{\Omega}P\,t+1}{\sum_{\Omega}P+\sum_{\Omega}t+1}.
  \end{aligned}
  \label{eq:loss}
\end{equation}
Since $g$ always gets a part in its own frame, one $g$ learns every kind of part at once. With several kinds of parts, each training crop comes from each kind with equal
probability and is cut at that kind's radius $R$; in our experiments, nails and teeth with probability $\frac12$ each.

\section{Experiments}
\label{sec:experiments}

The method rests on one idea: a small part is segmented best in its own frame, at the scanner's resolution. We test it
on two kinds of parts that differ in size and layout. A nail covers a tiny share of a hand scan and often faces away
from the scanner, while teeth fill much of an intraoral scan and sit close together. Both use the same settings; only
the part size $D$, which we measure from the labels, differs from one kind of part to the other.

\subsection{Setup}
\label{sec:setup}

\textbf{Data.}
We test on two datasets, one per kind of part.
\begin{itemize}[topsep=2pt,itemsep=3pt]
  \item \textbf{Nails.} 95 hand scans of 51 people: both hands of 44 people and one hand of the other 7. Each scan has
        about two million vertices, spaced 0.2--0.3\,mm apart. Human annotators outlined every nail by hand, and the
        vertices inside an outline take that nail's label. The annotators also rated each nail as usable or not; nails
        rated unusable are ignored in training and scoring. We split by person into 30 training people (57 hands, 213
        usable nails), 8 validation people (15 hands, 52 nails) and 13 test people (23 hands, 83 nails), and we score
        every usable test nail.
  \item \textbf{Teeth.} Teeth3DS+~\cite{benhamadou2022teeth3ds} intraoral scans (100--150k vertices, teeth labeled by
        their FDI number) with ToothGroupNetwork's patient folds~\cite{benhamadou2023teethseg}: 420 training, 90
        validation and 90 test patients, two jaws each. We also test on the 3DTeethSeg'22 private test set, the
        \emph{challenge test set}: 600 scans of 300 other patients, disjoint from all training scans.
\end{itemize}
Both networks are trained with augmentation (\cref{sec:app_impl}); validation and test scans are never augmented. The
segmentation network sees mirror images, smooth surface noise and the perturbed frames of \cref{sec:train}. The frame
network sees rotations, scalings, mirror images, and scans with smooth surface noise, holes and small tilts. The corrupted test scans of
\cref{tab:app_robust} are made separately: we decimate the test scans, add smooth noise to them, or cut one hole into
each nail (\cref{sec:robust}).

\textbf{Metrics.}
For nails, we report Dice per nail (mean and median over nails), the average symmetric surface distance (ASSD), the
boundary F-score at 0.25\,mm (BF$_{0.25}$) and the share of nails with Dice below 0.7. For teeth, we report the
per-tooth IoU, the boundary IoU~\cite{cheng2021boundary} at 0.25\,mm and ToothGroupNetwork's tooth identification rate
(TIR). The per-tooth IoU matches each ground-truth tooth to the predicted tooth overlapping it most, so a missed tooth
scores 0. Our models and the trained baselines use 3 seeds, and all tables report the mean over the seeds
(\cref{sec:app_data}).

\textbf{Baselines.}
Whole-scan methods on hands: Point Transformer V3 (PTv3)~\cite{wu2024ptv3}, MeshSegNet~\cite{lian2020meshsegnet} and
ToothGroupNetwork~\cite{benhamadou2023teethseg}, all retrained on our hands, and a \emph{whole-hand render}. It uses our
segmentation network, channels, 12 views within $60^\circ$ of the mat normal and lifting, but on the whole hand.
Per-part methods in our predicted frames: region growing~\cite{deng2026nail}, DGCNN~\cite{wang2019dgcnn} and PTv3 on the
crop. On teeth: the released TSegNet~\cite{cui2021tsegnet} and ToothGroupNetwork checkpoints;
ToothInstanceNet~\cite{vannistelrooij2025toothinstancenet} and the 2026 methods TeethSplineSeg~\cite{wei2026bspline}
and 3DTeethSAM~\cite{lu2026teethsam}, retrained on the training fold and selected on the validation fold; and a
\emph{whole-jaw render} like the whole-hand one. Every baseline trains on the same training scans as FrameSeg3D. \Cref{sec:app_baselines} gives details and
reproduction checks for every baseline.

\textbf{Implementation details.}
We use PyTorch, segmentation\_models\_pytorch, and Open3D ray casting. One segmentation network serves all kinds of
parts, in our experiments nails and teeth. Both networks train with AdamW~\cite{loshchilov2019adamw} and a one-cycle
schedule~\cite{smith2019superconvergence}, and we keep the checkpoint that is best on the validation set. All kinds of
parts share every setting; only $D$, which we measure from the labels, differs. We use a server with eight
NVIDIA H100 GPUs (80\,GB each), two Intel Xeon Platinum 8468 CPUs (128 cores), 2\,TB of memory and a 2\,TB disk;
each training run uses one GPU. \Cref{sec:app_impl} lists the hardware and software, the training schedules and times,
the data augmentation and the values of the remaining constants.

\subsection{Comparison with existing methods}
\label{sec:comparison}

\begin{table}[t]   %
\centering\small
\caption{\textbf{Nail segmentation} on the 83 test nails of 13 people (mean over 3 seeds). Dice: mean over nails;
ASSD: average symmetric surface distance in mm; BF$_{0.25}$: boundary F-score at 0.25\,mm; Fail: Dice below 0.7; render
pixel sizes in mm; GT: ground truth. Every baseline trains on the same training scans as FrameSeg3D. The whole-hand render uses our
network, channels, 12 views within $60^\circ$ and lifting, but on the whole hand. Bold marks the best value, and
underline the second best. The GT-frames row is not ranked.}
\label{tab:nails}
\setlength{\tabcolsep}{2.0pt}
\begin{tabularx}{\linewidth}{@{}lCCCCC@{}}
\toprule
Method & Dice$\uparrow$ & Dice$_{\mathrm{med}}\!\uparrow$ & ASSD$\downarrow$ & BF$_{0.25}\!\uparrow$ & Fail$\downarrow$ \\
\midrule
\multicolumn{6}{@{}l}{\emph{Whole scan}} \\
MeshSegNet~\cite{lian2020meshsegnet} & 0.735 & 0.797 & 1.010 & 0.198 & 22\% \\
ToothGroupNet.~\cite{benhamadou2023teethseg} & 0.797 & 0.874 & 0.633 & 0.374 & 18\% \\
PTv3~\cite{wu2024ptv3} & 0.953 & 0.958 & 0.201 & 0.675 & \underline{0.4\%} \\
Whole-hand render, 0.25\,mm & \underline{0.966} & 0.969 & 0.163 & 0.784 & \textbf{0\%} \\
Whole-hand render, 0.08\,mm & 0.956 & 0.968 & 0.168 & 0.777 & \underline{0.4\%} \\
\midrule
\multicolumn{6}{@{}l}{\emph{Per part, in our predicted frames}} \\
Region growing~\cite{deng2026nail} & 0.790 & 0.833 & 0.788 & 0.217 & 18\% \\
DGCNN~\cite{wang2019dgcnn} & \underline{0.966} & 0.969 & 0.160 & \underline{0.801} & \textbf{0\%} \\
PTv3~\cite{wu2024ptv3} & 0.961 & \underline{0.970} & \underline{0.153} & 0.794 & 1\% \\
\oursrow \textbf{FrameSeg3D (Ours)} & \textbf{0.970} & \textbf{0.973} & \textbf{0.144} & \textbf{0.829} & \textbf{0\%} \\
\oursrow FrameSeg3D (Ours), GT frames & 0.970 & 0.973 & 0.143 & 0.835 & 0\% \\
\bottomrule
\end{tabularx}
\end{table}

\begin{table}[t]   %
\centering\small
\caption{\textbf{Tooth segmentation on Teeth3DS+} with ToothGroupNetwork's patient folds. The challenge test set is the
3DTeethSeg'22 private test set (600 scans of 300 other patients). IoU: per-tooth IoU, each ground-truth tooth matched to
the predicted tooth it overlaps most (a missed tooth scores 0); BIoU: boundary IoU at 0.25\,mm. Both are means over
ground-truth teeth. TIR: predicted teeth with the right number, mean over scans. $\times100$; ours over 3 seeds.
TeethSplineSeg, 3DTeethSAM and ToothInstanceNet are retrained on the same training fold.}
\label{tab:teeth}
\setlength{\tabcolsep}{1.9pt}
\begin{tabularx}{\linewidth}{@{}lCCCCCC@{}}
\toprule
& \multicolumn{3}{c}{Test fold} & \multicolumn{3}{c}{Challenge set} \\
\cmidrule(lr){2-4}\cmidrule(l){5-7}
Method & IoU$\uparrow$ & BIoU$\uparrow$ & TIR$\uparrow$ & IoU$\uparrow$ & BIoU$\uparrow$ & TIR$\uparrow$ \\
\midrule
TSegNet~\cite{cui2021tsegnet} & 88.42 & 41.97 & 97.87 & 84.05 & 36.71 & 90.83 \\
ToothGroupNetwork~\cite{benhamadou2023teethseg} & \underline{95.96} & \underline{69.59} & \textbf{98.67} & 92.62 & 66.37 & 93.14 \\
TeethSplineSeg~\cite{wei2026bspline} & 94.86 & 60.26 & 98.26 & 92.32 & 59.60 & 90.72 \\
3DTeethSAM~\cite{lu2026teethsam} & 94.39 & 59.28 & 97.97 & 88.00 & 52.80 & 88.77 \\
ToothInstanceNet~\cite{vannistelrooij2025toothinstancenet} & 95.90 & 67.37 & \underline{98.63} & \textbf{96.23} & \underline{68.96} & \textbf{95.36} \\
Whole-jaw render, 0.08\,mm & 94.77 & 63.87 & 96.78 & 90.29 & 57.83 & 91.79 \\
\oursrow \textbf{FrameSeg3D (Ours)} & \textbf{96.48} & \textbf{70.46} & 98.41 & \underline{95.84} & \textbf{71.55} & \underline{94.05} \\
\bottomrule
\end{tabularx}
\end{table}

\begin{figure*}[t]
\centering
\includegraphics[width=\linewidth]{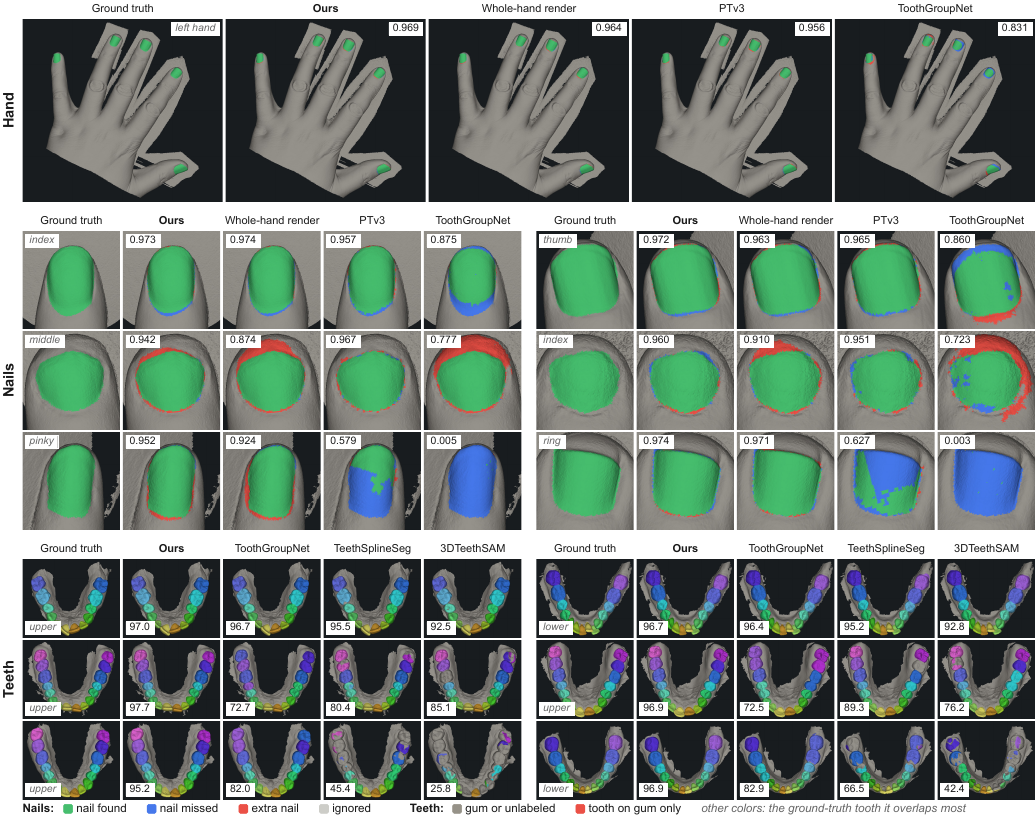}
\caption{\textbf{A typical test hand} (top, mean Dice over its five nails), \textbf{test nails} (middle, Dice per nail)
and \textbf{jaws from the challenge test set} (bottom, per-tooth IoU of the scan), picked automatically from the scores.
On the whole hand, nails are small and errors lie at their borders. Rows of nails and jaws: typical examples, where
every method is near its median; where the whole-hand render (nails) and ToothGroupNetwork (teeth) lose most against
FrameSeg3D; and where PTv3 (nails), TeethSplineSeg (teeth, left) and 3DTeethSAM (teeth, right) lose most. Nails are seen
along their normal, tip up. Each predicted tooth takes the color of the ground-truth tooth it overlaps most, so two
teeth in one color are one predicted tooth. More examples are in
\cref{fig:qual_hands_full,fig:qual_full,fig:qual_teeth_full}.}
\label{fig:qual}
\end{figure*}

\textbf{Nails.}
FrameSeg3D is the most accurate method on every metric (\cref{tab:nails}): mean Dice 0.970, ASSD 0.144\,mm,
BF$_{0.25}$ 0.829, and no failed nail. Every whole-scan method is behind: the whole-hand render by 0.004 mean Dice and
0.019\,mm ASSD, and PTv3 by 0.017. The two dental methods are 0.17 or more behind and fail on a fifth of the nails,
because they see too few points on a nail: ToothGroupNetwork keeps about 100 of its 24{,}000 points on a nail, and
MeshSegNet 41 of its 10{,}000 faces. Even perfect labels at MeshSegNet's input reach only 0.877. Our frames also help
other networks: PTv3 gains 0.008 mean Dice in our frames over the whole hand, and DGCNN in our frames comes within
0.004 of FrameSeg3D. Rendering in the frame still wins, most clearly on the boundary (BF$_{0.25}$ 0.829 vs.\ 0.801).
Predicted frames work as well as ground-truth ones (0.970 for both). \Cref{fig:qual} (top) shows typical baseline
failures: missed nail area (blue) and extra nail area (red), compared with the ground truth.

\textbf{Teeth.}
With the same settings as on nails and no dental knowledge, FrameSeg3D is the most accurate method on the test fold,
in per-tooth IoU (96.48) and in boundary IoU (70.46), ahead of ToothGroupNetwork (95.96 and 69.59), ToothInstanceNet
(95.90 and 67.37) and the two 2026 methods (\cref{tab:teeth}). On the challenge test set, it draws the most accurate
boundaries (71.55, against 68.96 for ToothInstanceNet) and leads ToothGroupNetwork, TeethSplineSeg and 3DTeethSAM by
3.2 to 7.8 per-tooth IoU. From the test fold to this unseen set, it loses only 0.6 per-tooth IoU, against 2.5 to 6.4
for these three methods. They typically merge neighboring teeth into one predicted tooth or leave teeth unlabeled
(\cref{fig:qual}, bottom). Only ToothInstanceNet has a higher per-tooth IoU on the challenge test set (96.23 vs.\
95.84). Its lead lies in the back molars, which are rare in our training fold (12 third molars in 840 scans, against
152 in the 600 challenge scans), so our frame network often places no frame on them. On incisors to first molars,
FrameSeg3D is ahead (96.52 vs.\ 96.34) and its boundary IoU is 2.7 higher (\cref{sec:app_teeth}).

\subsection{How a part is seen}
\label{sec:seeing}

\begin{table}[t]   %
\centering\small
\caption{\textbf{How a nail is seen.} The same network, channels and 12 views within $60^\circ$ of an axis. Only the
views' axis, center and pixel size change. Row 3 makes one pass, since re-centering would restore the frame normal. The
last row is FrameSeg3D. Pixels: median pixels on a nail per view. Test
nails, mean over 3 seeds.}
\label{tab:seeing}
\setlength{\tabcolsep}{2.4pt}
\begin{tabularx}{\linewidth}{@{}llCCCC@{}}
\toprule
Views around & Centered on & mm/px & Pixels & Dice$\uparrow$ & BF$_{0.25}\!\uparrow$ \\
\midrule
mat normal & whole hand & 0.25 & 1.2k & \underline{0.965} & 0.769 \\
mat normal & whole hand & 0.08 & 11.6k & 0.956 & 0.777 \\
mat normal & frame center & 0.08 & 11.5k & \underline{0.965} & \underline{0.801} \\
\oursrow \textbf{frame normal (Ours)} & frame center & 0.08 & 12.4k & \textbf{0.970} & \textbf{0.829} \\
\bottomrule
\end{tabularx}
\end{table}

\begin{table}[t]   %
\centering\small
\caption{\textbf{By nail tilt} (test nails, mean over 3 seeds; rows as in \cref{tab:seeing}). Tilt: angle between the
nail's normal and the mat normal; the number of nails in each bin in parentheses. Beyond $45^\circ$, views around the mat normal lose 0.038 against FrameSeg3D, and the
whole-hand render at 0.08\,mm/px loses 0.066 (0.899 vs.\ 0.965).}
\label{tab:seeing_tilt}
\setlength{\tabcolsep}{3pt}
\begin{tabularx}{\linewidth}{@{}llCCCCC@{}}
\toprule
& & & \multicolumn{4}{c}{Mean Dice$\uparrow$ by nail tilt from the mat} \\
\cmidrule(l){4-7}
Views around & Centered on & mm/px & \twoline{0--15$^\circ$\\(43)} & \twoline{15--30$^\circ$\\(24)} &
  \twoline{30--45$^\circ$\\(7)} & \twoline{${>}45^\circ$\\(9)} \\
\midrule
mat normal & whole hand & 0.25 & 0.970 & 0.963 & 0.956 & \underline{0.950} \\
mat normal & whole hand & 0.08 & 0.965 & 0.960 & 0.956 & 0.899 \\
mat normal & frame center & 0.08 & \underline{0.972} & \underline{0.966} & \underline{0.964} & 0.927 \\
\oursrow \textbf{frame normal (Ours)} & frame center & 0.08 & \textbf{0.973} & \textbf{0.967} & \textbf{0.967} & \textbf{0.965} \\
\bottomrule
\end{tabularx}
\end{table}

\begin{table}[t]   %
\centering\small
\caption{\textbf{Resolution} (test nails; whole-hand renders trained without augmentation). Inside the frame, 0.08 and
0.16\,mm nearly tie, and both beat 0.32\,mm, most clearly on the boundary. The whole hand gets worse with finer pixels.
At 0.08\,mm, the three whole-hand runs reach their best validation score after 1.5k, 4.5k and 5.5k of their 6k
training steps; we keep the checkpoint that is best on validation.}
\label{tab:app_res}
\setlength{\tabcolsep}{3pt}
\begin{tabularx}{\linewidth}{@{}lCCCCC@{}}
\toprule
Input & mm/px & Dice$\uparrow$ & Dice$_{\mathrm{med}}\!\uparrow$ & ASSD$\downarrow$ & BF$_{0.25}\!\uparrow$ \\
\midrule
Frame, 1 view, GT & 0.32 & 0.964 & 0.967 & 0.173 & 0.757 \\
 & 0.16 & 0.968 & 0.971 & 0.150 & 0.812 \\
 & 0.08 & 0.969 & 0.971 & 0.151 & 0.811 \\
Frame, 1 view, predicted & 0.32 & 0.964 & 0.968 & 0.170 & 0.766 \\
 & 0.16 & 0.968 & 0.970 & 0.157 & 0.797 \\
 & 0.08 & 0.969 & 0.971 & 0.152 & 0.807 \\
Whole hand, 12 views & 0.25 & 0.965 & 0.968 & 0.168 & 0.769 \\
 & 0.125 & 0.961 & 0.969 & 0.163 & 0.785 \\
 & 0.08 & 0.956 & 0.968 & 0.168 & 0.777 \\
\bottomrule
\end{tabularx}
\end{table}

We change only how a nail is seen and keep the network, channels, number of views and lifting fixed (\cref{tab:seeing}).
\emph{Pixel count does not explain the gap:} at the same 0.08\,mm per pixel, the whole-hand render puts almost as many
pixels on a nail as FrameSeg3D (11.6k vs.\ 12.4k per view), yet loses 0.014 mean Dice. \emph{Centering explains most
of it:} views around the mat normal, but centered on our frame, reach 0.965 with slightly fewer pixels on a nail
(11.5k). \emph{The view direction matters for tilted
parts} (\cref{tab:seeing_tilt}). Within $30^\circ$ of the mat, these views tie the views around the frame normal. Beyond
$45^\circ$, they fall to 0.927, against 0.965. Over all nails, they lose 0.028 BF$_{0.25}$. \emph{Resolution helps only
in the frame} (\cref{tab:app_res}). Inside the frame, finer pixels raise the mean Dice from 0.964 at 0.32\,mm to 0.969
at 0.08\,mm (0.968 at 0.16\,mm), and BF$_{0.25}$ from 0.757 to 0.811 with ground-truth frames. The whole-hand render instead gets worse, from 0.965 at 0.25\,mm to 0.956 at
0.08\,mm. \emph{Tilted parts gain most.} The margin over the whole-hand render grows from 0.003 for nails within
$15^\circ$ of the mat to 0.015 beyond $45^\circ$. There, the whole-hand render at 0.08\,mm falls to 0.899, while FrameSeg3D keeps a mean Dice of 0.965.

Teeth tilt less (median $14^\circ$ from the jaw axis). Here the frame helps mainly by spending the resolution on one
part. At the same 0.08\,mm, the frame adds 1.71 per-tooth IoU and 6.59 boundary IoU over the whole-jaw render. Viewing
along the tooth normal instead of the jaw axis adds only 0.15. Unlike the whole hand, the whole jaw does gain from finer
pixels: from 0.25 to 0.08\,mm/px it gains 0.88 per-tooth IoU and 7.73 boundary IoU (\cref{sec:app_teeth}).

\subsection{One network, few labels}
\label{sec:labels}

\begin{table}[t]   %
\centering\small
\caption{\textbf{One segmentation network for nails and teeth} (3 seeds each; Chall.: the challenge test set). A
dedicated network trains 6k steps on one domain. The shared network draws each crop's domain with probability
$\frac12$ and trains 24k steps, so each domain sees twice as many crops as with its dedicated network (12k steps' worth instead of 6k).}
\label{tab:app_joint}
\setlength{\tabcolsep}{2.6pt}
\begin{tabularx}{\linewidth}{@{}lCCCCC@{}}
\toprule
& \multicolumn{2}{c}{Nails} & \multicolumn{2}{c}{Teeth, test fold} & Chall. \\
\cmidrule(lr){2-3}\cmidrule(lr){4-5}\cmidrule(l){6-6}
Segmentation network & Dice$\uparrow$ & ASSD$\downarrow$ & IoU$\uparrow$ & BIoU$\uparrow$ & IoU$\uparrow$ \\
\midrule
Two dedicated networks & 0.970 & 0.142 & 96.46 & 70.34 & 95.73 \\
\oursrow One network, 24k steps & 0.970 & 0.144 & 96.48 & 70.46 & 95.84 \\
\bottomrule
\end{tabularx}
\end{table}

\begin{table}[t]   %
\centering\small
\caption{\textbf{Few labels on nails.} Every method trains on the same nested subsets of the 30 training people.
\emph{Segmentation network on the subset}: frames from the frame network trained on all data. \emph{Both stages}: the
frame network is also trained on the subset. Subsets: one seed. All 30 people: mean over 3 seeds, as in
\cref{tab:nails}. Each row keeps one training recipe at every subset size, so its all-data value can differ slightly
from \cref{tab:nails}.}
\label{tab:app_labels}
\setlength{\tabcolsep}{2.8pt}
\begin{tabularx}{\linewidth}{@{}lCCCC@{}}
\toprule
& \multicolumn{4}{c}{Mean Dice$\uparrow$ on test nails} \\
\cmidrule(l){2-5}
Training people (of 30) & 1 & 3 & 8 & 30 \\
\midrule
\oursrow Ours, segmentation network on the subset & \textbf{0.949} & \textbf{0.966} & \textbf{0.968} & \textbf{0.970} \\
\oursrow Ours, both stages on the subset & 0.772 & 0.943 & \underline{0.967} & \textbf{0.970} \\
DGCNN in our frames, segmentation on the subset & \underline{0.935} & 0.956 & 0.964 & \underline{0.966} \\
DGCNN in our frames, both stages on the subset & 0.747 & 0.930 & 0.964 & \underline{0.966} \\
Whole-hand render & 0.888 & \underline{0.959} & 0.961 & 0.964 \\
PTv3, whole hand & 0.568 & 0.901 & 0.932 & 0.951 \\
\bottomrule
\end{tabularx}
\end{table}

\begin{table}[t]   %
\centering\small
\caption{\textbf{Few labels on teeth.} Every method trains on the same nested subsets of the 420 training patients.
\emph{Segmentation network on the subset} and \emph{both stages} as in \cref{tab:app_labels}. Subsets: one seed. All 420
patients: mean over 3 seeds, as in \cref{tab:teeth}. ToothGroupNetwork is retrained with its official code on the
subsets; with all 420 patients it is the released checkpoint.}
\label{tab:app_labels_teeth}
\setlength{\tabcolsep}{2.8pt}
\begin{tabularx}{\linewidth}{@{}lCCC@{}}
\toprule
& \multicolumn{3}{c}{Per-tooth IoU$\uparrow$ on the test fold} \\
\cmidrule(l){2-4}
Training patients (of 420) & 5 & 42 & 420 \\
\midrule
\oursrow Ours, segmentation network on the subset & \textbf{96.38} & \textbf{96.41} & \textbf{96.48} \\
\oursrow Ours, both stages on the subset & \underline{96.16} & \underline{96.36} & \textbf{96.48} \\
Whole-jaw render, 0.08\,mm/px & 92.36 & 94.23 & 94.77 \\
ToothGroupNetwork & 87.18 & 89.40 & \underline{95.96} \\
\bottomrule
\end{tabularx}
\end{table}

\textbf{One network for nails and teeth.}
We train one segmentation network on crops of both kinds, drawn with equal probability. It matches the two dedicated
networks (\cref{tab:app_joint}): mean Dice 0.970 vs.\ 0.970 on nails, per-tooth IoU 96.48 vs.\ 96.46 on the test fold, and 95.84 vs.\ 95.73
on the challenge test set. It trains for 24{,}000 steps, twice as many per domain as a dedicated network. Initializing one
domain from the other's network does not help (\cref{sec:app_labels}). The two kinds of parts share a network but do not
help each other.

\textbf{Few labels.}
In its canonical view, the segmentation network needs few masks (\cref{tab:app_labels,tab:app_labels_teeth}). We first use frames from the full
frame network. Trained
on the nails of 3 of the 30 people, the segmentation network already beats the whole-hand render trained on all 30
(0.966 vs.\ 0.964). With one person, it leads the render trained on that person by 0.061 (0.949 vs.\ 0.888). It also needs fewer labels than a point network
in the same frames: with 1, 3 and 8 people, it leads DGCNN in our frames by 0.014, 0.010 and 0.004 mean Dice. Without a
frame, PTv3 on the whole hand falls to 0.568 with one person. On teeth (\cref{tab:app_labels_teeth}),
cutting the training patients from 420 to 42 or 5 costs it 0.07 or 0.10 per-tooth IoU, but costs the whole-jaw render
0.54 or 2.41. ToothGroupNetwork, retrained with its own code on the same 42 or 5 patients, falls to 89.40 or 87.18,
from 95.96 for its released checkpoint on all 420. With 5 patients, our segmentation network still beats that
checkpoint (96.38 vs.\ 95.96). The frame network is the weaker stage.
Trained on the subset too, it loses two nails at 3 people (mean Dice 0.943), and at one person it finds only 81\% of
the nails (0.772). Still, the whole method needs only a fraction of the labels. With both stages trained on 8 of the 30 people,
FrameSeg3D beats the whole-hand render trained on all 30 (0.967 vs.\ 0.964). With both stages trained on 5 of 420
patients, FrameSeg3D leads ToothGroupNetwork trained on the same patients by 8.98 per-tooth IoU (96.16 vs.\ 87.18)
and still beats it trained on all 420 (96.16 vs.\ 95.96). With 42 patients, a tenth of the training fold, it reaches
96.36, within 0.12 of training on all 420.

\subsection{Ablations}
\label{sec:ablations}

\begin{table}[t]   %
\centering\small
\caption{\textbf{Ablations.} One part of FrameSeg3D changed at a time (test, mean over 3 seeds, all at 0.08\,mm per
pixel). No frame: the whole-hand and whole-jaw renders. $^{\dagger}$Centered on the part; nails with predicted centers
and one pass, teeth with ground-truth centers. $^{\ddagger}$Teeth with ground-truth frames; with them, 12 views reach 96.53 per-tooth IoU.}
\label{tab:ablation}
\setlength{\tabcolsep}{2.6pt}
\begin{tabularx}{\linewidth}{@{}lCCCC@{}}
\toprule
& \multicolumn{2}{c}{Nails} & \multicolumn{2}{c}{Teeth, test fold} \\
\cmidrule(lr){2-3}\cmidrule(l){4-5}
Variant & Dice$\uparrow$ & BF$_{0.25}\!\uparrow$ & IoU$\uparrow$ & BIoU$\uparrow$ \\
\midrule
\oursrow \textbf{FrameSeg3D (Ours)} & 0.970 & 0.829 & 96.48 & 70.46 \\
\midrule
No frame: views of the whole scan & 0.956 & 0.777 & 94.77 & 63.87 \\
Views around the scan's up axis$^{\dagger}$ & 0.965 & 0.801 & 96.38 & 69.67 \\
One view along the frame normal$^{\ddagger}$ & 0.968 & 0.794 & 86.63 & 39.68 \\
No re-centering & 0.970 & 0.832 & 96.45 & 70.40 \\
Ground-truth frames & 0.970 & 0.835 & 96.53 & 70.50 \\
Two dedicated segmentation networks & 0.970 & 0.836 & 96.46 & 70.34 \\
\bottomrule
\end{tabularx}
\end{table}

\Cref{tab:ablation} changes one part of FrameSeg3D at a time. Without a frame, both tasks lose clearly (nails $-0.014$
Dice, teeth $-1.71$ IoU). Views around the scan's up axis lose mainly at the boundary, and one view fails on teeth.
Re-centering, ground-truth frames and two dedicated networks change little, so predicted frames and one shared
network are enough.

\begin{table}[t]
\setlength{\linewidth}{0.64\textwidth}   %
\centering\small
\captionsetup{width=0.64\textwidth}   %
\caption{\textbf{Frame robustness} (median / mean Dice over validation and test nails, 3 seeds). The ground-truth frame
is tilted or shifted in a random direction. One pass, or one re-centering pass that recomputes the frame from the first
segment by \cref{eq:frame}.}
\label{tab:app_frame}
\setlength{\tabcolsep}{7pt}
\begin{tabular}{@{}lcc@{}}
\toprule
& \multicolumn{2}{c}{Median / mean Dice$\uparrow$} \\
\cmidrule(l){2-3}
Ground-truth frame perturbed by & One pass & Re-centered \\
\midrule
none & 0.973 / 0.969 & 0.972 / 0.969 \\
tilt $20^\circ$ & 0.972 / 0.968 & 0.973 / 0.969 \\
tilt $45^\circ$ & 0.968 / 0.959 & 0.973 / 0.969 \\
shift 4\,mm & 0.972 / 0.969 & 0.973 / 0.969 \\
shift 8\,mm & 0.946 / 0.837 & 0.972 / 0.913 \\
\bottomrule
\end{tabular}
\end{table}

\textbf{Frames.}
The frame network names every test finger correctly (median center error 0.38\,mm, normal error $4.0^\circ$). On teeth,
it places a frame with the right number within 3\,mm of 99.6\% of the test teeth (center 0.52\,mm, normal
$5.4^\circ$). The segmentation network tolerates such errors. Its result does not change when the frame is tilted by $20^\circ$ or shifted by 4\,mm. Re-centering recovers a
$45^\circ$ tilt (mean Dice 0.959 to 0.969; \cref{tab:app_frame}). Only a shift of 8\,mm, about $0.7D$, hurts (mean
Dice 0.837, and 0.913 after re-centering). With predicted frames, re-centering changes the mean Dice on nails by less
than 0.001, so it matters only for large frame errors.

\textbf{Views.}
One view along the frame normal loses only 0.002 mean Dice on nails but 0.035 BF$_{0.25}$ (0.794 vs.\ 0.829). It fails
on teeth (per-tooth IoU 86.6 vs.\ 96.5 with ground-truth frames). A view from above does not see the side walls of a
crown, where the gum line runs. The gain comes from the views at test time. A cone-trained network tested on one view is no better than one trained on that view, so the cone helps only when it is also used at test time.

\textbf{Scan quality and cost.}
FrameSeg3D keeps the highest mean Dice on test hands that are decimated, made noisy or cut by holes
(\cref{tab:app_robust}). Its mean Dice stays at 0.964 or above under every corruption we test. The whole-scan models
fail where their input changes most: PTv3 falls to 0.461 with an eighth of the faces, and the whole-hand render to 0.891
under 0.1\,mm of smooth noise. FrameSeg3D takes 15.5\,s per hand on one GPU, half the time of the whole-hand render at
the same 0.08\,mm per pixel. Most of this time goes to the in-frame segmentation, which renders and segments 12 views of
each part in two passes (\cref{sec:robust}).

\section{Conclusion}
\label{sec:conclusion}
Small parts on high-resolution 3D scans are limited by the direction and scale at which they are seen, not by how many
samples land on them. FrameSeg3D turns this into a simple method. It predicts a frame for every part from a coarse view of
the whole scan, then segments each part in views around its frame normal at the scanner's resolution. With the same
settings, it outperforms whole-scan and two-stage methods on nails and on the Teeth3DS+ test fold, and it draws the
most accurate tooth boundaries. On hands, the gain over the whole-hand render grows with the nail's tilt and is largest on
the thumb, which faces sideways. The frames also help other networks: in our frames, DGCNN comes within 0.004 mean
Dice of FrameSeg3D on nails. The canonical view also lets one segmentation network serve both kinds of parts and
learn from a fraction of the labels: a quarter of the hands or a tenth of the patients. Between nails and teeth, only
the part size changes, and we measure it from the labels. So we expect a new kind of part to need labeled scans, not a
new design.

\section*{Limitations}
The frame network is the weak stage. It needs more labels than the segmentation network: trained on the hands of one
person, it misses a fifth of the nails. It misses parts that are rare in training: on the challenge test set, it places no
frame on many third molars, which the training fold almost lacks. There, ToothInstanceNet segments and names teeth more
accurately. The frame network also assumes that each part faces a clear direction. On jaws, where the crops of neighboring teeth overlap,
FrameSeg3D is slower than rendering the whole jaw and than the dental methods (21.9\,s per jaw, against 1.3 to
8.8\,s). We test only two kinds of parts, and all hand scans come from one 3D scanner. We simulate lower scan quality by
decimation, noise and holes, but we have not tested scans from other scanners. Learning the
frame without frame labels, for example from the segmentation loss through a differentiable renderer, is a natural next
step, since the two stages could then be trained together, end to end.

\section*{LLM Usage Statement}
The idea was proposed and the initial experiments were conducted by the human authors. Claude Code (Anthropic) and
Codex (OpenAI) helped improve the experiments and analyze the data. They also helped polish and revise the text. The
human authors take full responsibility for all ideas, experiments, conclusions, claims and contributions, and for all of
the text.

\clearpage
{
    \small
    \bibliographystyle{ieeenat_fullname}
    \bibliography{main}
}

\clearpage
\maketitlesupplementary   %
\appendix
\crefalias{section}{appendix}   %

\section{Data and metrics}
\label{sec:app_data}

\textbf{Hand scans.}
Each hand lies flat on a mat and is scanned with one 3D scanner (about two million vertices, 0.2--0.3\,mm apart). Human
annotators outlined each nail by hand and labeled every vertex as nail (with the finger's name), skin, or
\emph{ignore} (doubled scan layers and loose fragments). Ignored vertices are excluded from scoring and from all
losses. The annotators also rated each nail as usable or not. Seven people have one hand in the dataset (three training, one
validation and three test people), and the others both hands. The 83 test nails
(11 thumbs) are those with labels rated usable. The split is by person: 30 training, 8 validation and 13 test people, with 57, 15 and 23 hands and 213, 52 and 83
usable nails.

\textbf{Nail metrics.}
Each ground-truth nail is scored against the predicted nail pieces that touch it, whatever their finger label, so finger
names do not affect the scores. Dice weights each vertex by its area. ASSD is the mean distance between the predicted
and ground-truth outlines, in both directions. BF$_{0.25}$ is the F-score of outline vertices within 0.25\,mm of the
other outline: precision counts the predicted outline vertices near the true outline, and recall the reverse.

\textbf{Tooth metrics.}
ToothGroupNetwork's evaluation code averages IoU and F1 over \emph{predicted} teeth, so a missed tooth does not count.
Its TIR is the share of predicted teeth with the right number. So we also report our per-tooth IoU. It matches each
ground-truth tooth to the predicted tooth that overlaps it most, ignoring tooth numbers. A missed tooth scores 0.
Boundary IoU~\cite{cheng2021boundary} is computed on the mesh. We score all teeth results with one scorer, from
per-vertex predictions on the same meshes. Per-tooth IoU and boundary IoU are means over ground-truth teeth.
ToothGroupNetwork's metrics are means over scans, as in its evaluation code.

\textbf{Seeds.}
Our models and the trained baselines use 3 seeds; tables give the mean. The segmentation networks
trained on label subsets (\cref{sec:app_labels}) use one seed. A frame network on a subset is still an ensemble of 3, as
in the full method, with each member trained on the same subset from a different random seed, and their outputs are averaged before finding peaks.

\section{Scan quality and cost}
\label{sec:robust}

\begin{table*}[!htbp]
\centering\small   %
\caption{\textbf{Robustness to scan quality} (test nails, mean Dice, 3 seeds). We decimate the test scans, displace them
by smooth noise along the normal (correlation length about 0.6\,mm), or cut one hole per nail and close it with flat
triangles. Every model sees surface noise in training, and the whole-scan models and our frame network also see
holes.
The whole-hand render uses 0.25\,mm per pixel. FrameSeg3D is best under every corruption and at every strength.}
\label{tab:app_robust}
\setlength{\tabcolsep}{3pt}   %
\begin{tabularx}{\linewidth}{@{}lCCCCCCCCCC@{}}
\toprule
\multicolumn{2}{@{}l}{Mean Dice$\uparrow$} & \multicolumn{3}{c}{Faces kept} & \multicolumn{3}{c}{Smooth noise (mm)} & \multicolumn{3}{c}{Hole radius (mm)} \\
\cmidrule(lr){3-5}\cmidrule(lr){6-8}\cmidrule(l){9-11}
Method & Clean & 1/2 & 1/4 & 1/8 & 0.025 & 0.05 & 0.1 & 1 & 2 & 3 \\
\midrule
\oursrow \textbf{FrameSeg3D (Ours)} & \textbf{0.970} & \textbf{0.969} & \textbf{0.967} & \textbf{0.964} & \textbf{0.969} & \textbf{0.966} &
  \textbf{0.964} & \textbf{0.970} & \textbf{0.968} & \textbf{0.964} \\
DGCNN in the frame & \underline{0.966} & \underline{0.964} & \underline{0.960} & \underline{0.955} & \underline{0.965} & \underline{0.962} & \underline{0.954} & \underline{0.966} & \underline{0.964} & \underline{0.961} \\
Whole-hand render & \underline{0.966} & 0.961 & 0.957 & 0.952 & 0.964 & \underline{0.962} & 0.891 & 0.965 & \underline{0.964} & 0.960 \\
PTv3, whole hand & 0.953 & 0.950 & 0.918 & 0.461 & 0.952 & 0.948 & 0.932 & 0.951 & 0.950 & 0.943 \\
\bottomrule
\end{tabularx}
\end{table*}

\textbf{Scan quality.}
We corrupt the test scans in three ways: we decimate them, displace them by smooth noise along the normal, or cut one
hole into each nail and close it with flat triangles (\cref{tab:app_robust}). FrameSeg3D is best under every density,
noise level and hole size, and its mean Dice never falls below 0.964, also with an eighth of the faces, under 0.1\,mm
of smooth noise and with a 3\,mm hole on every nail. The whole-scan models fail where their input changes most: PTv3
falls to 0.461 with an eighth of the faces, and the whole-hand render to 0.891 under 0.1\,mm of noise.

\begin{table}[!htbp]   %
\centering\small
\caption{\textbf{Inference cost} with the final networks on one H100 GPU (16 CPU threads), mean over six test scans
after one warm-up. Time: from the loaded mesh to per-vertex labels, without reading or writing files; for FrameSeg3D and
the renders, the vertex normals and curvature are loaded with the mesh. GPU memory: peak PyTorch allocation / peak use of
the process, which adds the CUDA context and the allocator's cache; for the two stages of FrameSeg3D, the peak
allocation while that stage runs, with the networks of both stages loaded. FrameSeg3D's parameters are the three frame networks
of the ensemble (24.4\,M each) and one segmentation network (24.4\,M).}
\label{tab:app_cost}
\setlength{\tabcolsep}{3pt}
\begin{tabularx}{\linewidth}{@{}lCCC@{}}
\toprule
Method & Time (s)$\downarrow$ & GPU memory (GB)$\downarrow$ & Parameters (M)$\downarrow$ \\
\midrule
\multicolumn{4}{@{}l}{\emph{Hands, per hand (1.6--3.3 million vertices)}} \\
\oursrow \textbf{FrameSeg3D (Ours)} & 15.5 & 1.3 / 2.6 & 97.8 \\
\oursrow \quad frame prediction & 2.5 & 0.7 & 73.3 \\
\oursrow \quad in-frame segmentation, two passes & 12.9 & 1.3 & 24.4 \\
\oursrow FrameSeg3D (Ours), one view per frame & 9.3 & 0.7 / 1.5 & 97.8 \\
DGCNN in the frame & 6.6 & 7.4 / 28.3 & 74.3 \\
Whole-hand render, 0.25\,mm/px & 6.7 & 0.5 / 1.3 & 24.4 \\
Whole-hand render, 0.08\,mm/px & 30.6 & 3.9 / 6.4 & 24.4 \\
PTv3, whole hand & 0.9 & 1.5 / 3.7 & 46.2 \\
\midrule
\multicolumn{4}{@{}l}{\emph{Teeth, per jaw (78--176k vertices)}} \\
\oursrow \textbf{FrameSeg3D (Ours)} & 21.9 & 1.3 / 2.3 & 97.8 \\
\oursrow \quad frame prediction & 0.2 & 0.4 & 73.3 \\
\oursrow \quad in-frame segmentation, two passes & 21.7 & 1.3 & 24.4 \\
Whole-jaw render, 0.25\,mm/px & 2.4 & 0.2 / 0.8 & 24.4 \\
Whole-jaw render, 0.08\,mm/px & 8.8 & 0.9 / 2.0 & 24.4 \\
TSegNet & 1.3 & 1.7 / 3.7 & 2.4 \\
ToothGroupNetwork & 4.4 & 1.3 / 4.4 & 15.8 \\
ToothInstanceNet & 2.0 & 4.1 / 13.0 & 27.1 \\
TeethSplineSeg & 6.9 & 0.3 / 1.5 & 47.4 \\
3DTeethSAM & 6.1 & 5.2 / 9.7 & 296.9 \\
\bottomrule
\end{tabularx}
\end{table}

\textbf{Cost.}
\Cref{tab:app_cost} gives the inference cost of every method with the final networks, measured on one GPU in the same
way. On hands, FrameSeg3D takes 15.5\,s per hand, half the time of the whole-hand render at the same 0.08\,mm per pixel
(30.6\,s), with a third of its allocated GPU memory. The whole-hand render at 0.25\,mm and PTv3 are faster but less
accurate (\cref{tab:nails}). The GPU memory of FrameSeg3D stays small on hands and on jaws, 1.3\,GB of allocations,
because every view has the same $384\times384$ pixels whatever the size of the scan. On teeth, it is the slowest method
(21.9\,s per jaw, about 1.5\,s per tooth), because the crops of neighboring teeth overlap and are rendered one at a
time; batching them would make it faster. Most of its parameters belong to the three frame networks of the ensemble.

\FloatBarrier
\section{Implementation details}
\label{sec:app_impl}

\textbf{Hardware and software.}
We use a server with eight NVIDIA H100 GPUs with 80\,GB each, two Intel Xeon Platinum 8468 CPUs with 128 cores in
total, 2\,TB of memory, and a 2\,TB disk for the data and caches. Our code uses Python 3.10, PyTorch 2.5.1 with CUDA
12.4, segmentation\_models\_pytorch, and Open3D for ray casting; the baselines run in their own environments. Each
training run uses one GPU, and several runs share the server at a time. With other runs on the server, one seed of the
shared segmentation network (24k steps) trains in about 6.4\,h on one GPU, and one seed of the frame network (20k
steps) in about 2.7\,h on hands and 0.6\,h on teeth. \Cref{sec:robust} gives the inference cost of every method.

\textbf{Overview image.}
The image spans 1.2 times the 90th percentile of the training scans' extent, so that nearly every scan fits with a
margin. We find the local floor by a grayscale opening of the height image over $12D$, much wider than a part, so
that the opening removes the parts and keeps what they rest on; the overview's height channel is measured from this floor.

\textbf{Frame network.}
The heatmap Gaussians have a standard deviation of $D/8$. Twice that is $D/4$, well below the distance between the
centers of neighboring parts, so their peaks stay apart. The normal loss uses the pixels within $D/6$ of a part's center
($w(\mathbf{q})=1$ in \cref{eq:frameloss}), where the surface faces most like the part as a whole. Pixels with an ignore
label give no loss, and pixels within $D/2$ of one give no heatmap loss, because an ignored part still looks like a
part. A part is a local maximum of $\hat C$ above 0.3 that lies at least $D/3$ from a stronger one; $D/3$ is well below
the distance between neighboring parts. The Hungarian algorithm scores each name by the mean of $\hat{\mathbf{S}}$
within $D/6$ of a peak. The network never sees scan tags, such as whether a hand is left or right or a jaw upper or
lower; a tag only maps a predicted name to the dataset's label, for instance a tooth position, counted from the front
of the jaw, to its tooth number in the upper or lower jaw. We set these constants once and use them for every kind of
part.

\textbf{Rendering.}
We divide the height channel by 5\,mm and clip the curvature to $[-1,1]$ to keep the channels in a similar range as the
normal components, which lie in $[-1,1]$.

\textbf{Training schedules.}
The segmentation network trains for 24{,}000 steps of 16 crops (peak learning rate $5\times10^{-4}$), so that each kind
of part gets twice the steps of a network trained on one kind (6{,}000). We keep the checkpoint with the best mean,
over the kinds of parts, of the median crop Dice on up to 300 validation parts each. The frame network trains for
20{,}000 steps of 16 overview images (learning rate $5\times10^{-4}$); at 6{,}000 steps its validation detection score
was still rising. We keep its checkpoint with the best detection F1 on validation, with the right name within $D/3$.

\textbf{Data augmentation.}
We augment training scans only. Each training crop of the segmentation network is mirrored with probability $\frac12$.
With probability $\frac12$, we also move it along its normals by a smooth field $\eta$,
$\mathbf{v}_i\leftarrow\mathbf{v}_i+\eta_i\mathbf{n}_i$: white noise smoothed over the mesh graph (correlation length about
0.6\,mm) with a robust standard deviation drawn from $\mathcal{U}(0,0.12\,\mathrm{mm})$. We then recompute normals and
curvature, so the network cannot rely on the noise level of one scanner. The frame network sees any rotation, scaling
by 0.9--1.1 and mirroring. A mirror image renames the parts by a fixed map: a tooth position changes sides, while a
finger keeps its name. Half of its training images come from one of eight augmented copies of each training scan.
Each copy has smooth surface noise as above, a hole on each part with probability $\frac12$, and a tilt of up to
$15^\circ$ about a random axis perpendicular to $\mathbf{u}$. A hole is a ball of radius up to $0.3D$ around a random
point of the part, cut out and closed with flat triangles, much like Cutout~\cite{devries2017cutout} on a surface. We fixed these settings in advance and did not tune them. The
whole-scan baselines are trained with the same noise and holes (\cref{sec:app_baselines}).

\section{Baselines}
\label{sec:app_baselines}

\begin{table*}[!htbp]
\centering\small   %
\caption{\textbf{All methods on the Teeth3DS+ test fold} (180 scans of 90 patients, $\times100$). Their metrics:
ToothGroupNetwork's IoU, F1 and TIR, averaged over predicted teeth and then over scans. Per ground-truth tooth: our
per-tooth IoU and boundary IoU, means over ground-truth teeth. Jaw-axis views: views around the jaw axis instead of the
tooth normal. Ours and the whole-jaw renders: 3 seeds. Retrained methods: 1 run each. With ground-truth frames or
centers, TIR is 100 by construction, since the frames carry the tooth numbers. These rows are not ranked.}
\label{tab:app_teeth}
\setlength{\tabcolsep}{3pt}
\begin{tabularx}{\linewidth}{@{}lCCCCcc@{}}
\toprule
& \multicolumn{3}{c}{Their metrics} & \multicolumn{3}{c}{Per ground-truth tooth} \\
\cmidrule(lr){2-4}\cmidrule(l){5-7}
Method & IoU$\uparrow$ & F1$\uparrow$ & TIR$\uparrow$ & IoU$\uparrow$ & BIoU$_{0.25}\!\uparrow$ & BIoU$_{0.5}\!\uparrow$ \\
\midrule
PointNet & 78.38 & 85.89 & 92.63 & 81.12 & 29.26 & 38.32 \\
DGCNN (released checkpoint) & 83.27 & 89.08 & 94.02 & 86.24 & 33.91 & 43.16 \\
PointNet++ & 89.54 & 93.56 & 96.58 & 91.26 & 44.25 & 56.51 \\
TSegNet~\cite{cui2021tsegnet} & 90.77 & 94.83 & 97.87 & 88.42 & 41.97 & 54.02 \\
Point Transformer & 91.09 & 94.64 & 97.08 & 92.31 & 48.70 & 61.00 \\
ToothGroupNetwork~\cite{benhamadou2023teethseg} & \underline{95.99} & \textbf{97.83} & \textbf{98.67} & \underline{95.96} & \underline{69.59} & \underline{78.08} \\
\quad without boundary-aware sampling & 93.00 & 96.25 & \textbf{98.67} & 92.97 & 50.55 & 62.92 \\
TeethSplineSeg~\cite{wei2026bspline}, retrained & 94.64 & 97.03 & 98.26 & 94.86 & 60.26 & 71.24 \\
3DTeethSAM~\cite{lu2026teethsam}, retrained & 93.58 & 96.17 & 97.97 & 94.39 & 59.28 & 69.85 \\
ToothInstanceNet~\cite{vannistelrooij2025toothinstancenet}, retrained & 95.63 & 97.58 & \underline{98.63} & 95.90 & 67.37 & 76.71 \\
Whole-jaw render, 0.25\,mm/px & 92.84 & 95.66 & 97.80 & 93.89 & 56.14 & 67.64 \\
Whole-jaw render, 0.08\,mm/px & 92.99 & 95.28 & 96.78 & 94.77 & 63.87 & 73.46 \\
\oursrow \textbf{FrameSeg3D (Ours)} & \textbf{96.11} & \underline{97.82} & 98.41 & \textbf{96.48} & \textbf{70.46} & \textbf{79.03} \\
\oursrow \quad ground-truth frames & 96.53 & 98.21 & 100.00 & 96.53 & 70.50 & 79.08 \\
\oursrow \quad ground-truth centers, jaw-axis views & 96.37 & 98.12 & 100.00 & 96.38 & 69.67 & 78.25 \\
\oursrow \quad one view, ground-truth frames & 86.61 & 92.43 & 100.00 & 86.63 & 39.68 & 44.44 \\
\bottomrule
\end{tabularx}
\end{table*}

\textbf{On hands.}
All hand baselines train on the same 30 people and select checkpoints on the same 8.
\emph{PTv3}~\cite{wu2024ptv3} segments nail vs.\ not nail on the hand above the mat, at its 0.5\,mm grid (median 671
points per nail). We drop nail pieces under 20\,mm$^2$.
\emph{MeshSegNet}~\cite{lian2020meshsegnet} sees the hand decimated to about 10{,}000 faces by vertex clustering,
because quadric decimation stalls on most hands. Its output is upsampled to every vertex by nearest neighbors. We could
not run its graph-cut refinement. Its input resolution limits it: even perfect labels at that input reach a mean Dice of
only 0.877.
\emph{ToothGroupNetwork}~\cite{benhamadou2023teethseg} samples 24{,}000 points (median 98 per nail). For hands, we use
full rotation augmentation, a fixed 70\,mm scale, and a label vote per cluster instead of tooth numbers. We select its
DBSCAN parameter on validation and run it three times with a time-capped schedule.
The \emph{whole-hand render} runs our segmentation network, channels and lifting on the whole hand, with 12 Fibonacci
views within $60^\circ$ of the mat normal and 0.25, 0.125 or 0.08\,mm per pixel. Each image covers the hand.
\emph{DGCNN}~\cite{wang2019dgcnn} and \emph{PTv3} in the frame segment the same crops as FrameSeg3D, trained on the same
perturbed frames. \emph{Region growing}~\cite{deng2026nail} grows the nail from the frame center over the surface.
All baselines train on the same training scans as FrameSeg3D. DGCNN and PTv3 in the frame train with our surface
noise, like FrameSeg3D. PTv3 and the whole-hand render at 0.25\,mm per pixel see surface noise and holes like those of
the frame network
(\cref{sec:app_impl}), drawn from 16 precomputed augmented versions per hand, because per-step noise was too slow to
compute on whole hands of two million vertices.

\textbf{On teeth.}
We run the released checkpoints of ToothGroupNetwork and its baselines on its own test fold. They match the published
table within 0.01 IoU for ToothGroupNetwork (with and without its boundary-aware sampling), Point Transformer and
PointNet. TSegNet is 1.1 below and DGCNN 5.0 below. We retrain
\emph{TeethSplineSeg}~\cite{wei2026bspline} from scratch on the training fold with its full pipeline and
post-processing, and select on the validation fold. Its own evaluation on our test fold gives 93.55 T-mIoU and 70.87
B-IoU, close to its paper's 92.62 and 71.36. We retrain \emph{3DTeethSAM}~\cite{lu2026teethsam} from SAM2.1 Hiera-L with
its two-stage schedule (30 + 70 epochs), seven views and graph-cut smoothing. Its own evaluation gives 93.44 T-mIoU,
against 91.90 in its paper. Its released checkpoint saw 120 of our 180 test scans in training, so we do not use it. We
retrain \emph{ToothInstanceNet}~\cite{vannistelrooij2025toothinstancenet} from scratch with its released code and full
inference pipeline (pose alignment, tooth instances, then FDI numbers). Its released checkpoints were trained on
Teeth3DS scans that may include our test scans, so we do not use them. Of the two training configurations in its
repository, we keep the one with the higher validation F1. The \emph{whole-jaw render} predicts 17 classes (background
and 16 positions) per pixel, because neighboring teeth touch. Each tooth keeps its largest piece, as in FrameSeg3D. We found no public code for Render2Seg~\cite{lee2025render2seg}, TSegLab~\cite{rekik2025tseglab} or
IOSSAM~\cite{huang2024iossam}, so we do not compare with them.

\textbf{All teeth methods.}
\Cref{tab:app_teeth} lists every method on the test fold, with ToothGroupNetwork's own metrics next to ours. Their IoU
and F1 average over predicted teeth and then over scans, so they do not count a missed tooth; our per-tooth IoU does.
With ground-truth frames, the tooth numbers come from the ground truth, so TIR is 100 by construction.

\FloatBarrier
\section{More results on nails}
\label{sec:app_nails}

\begin{figure*}[!htb]   %
\centering
\includegraphics[width=\linewidth]{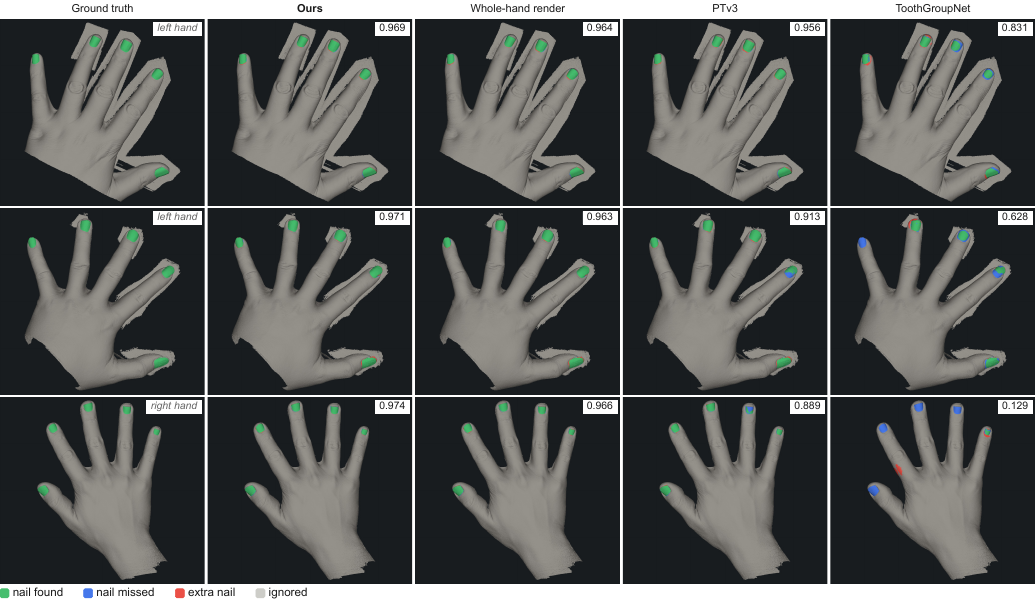}
\caption{\textbf{Whole test hands} seen from above, with each method's mean Dice over the five nails. The hands are
picked automatically: the typical hand of \cref{fig:qual}, the hand where the whole-hand render loses most against
FrameSeg3D, and the hand where both PTv3 and ToothGroupNetwork lose most. ToothGroupNetwork misses most nails of this
hand. We show only hands with all five nails scored. Nails with unusable labels are ignored. Detached scan fragments and
the board under one hand are not drawn, so that each tile shows only the hand.}
\label{fig:qual_hands_full}
\end{figure*}

\begin{table}[!htbp]   %
\centering\small
\caption{\textbf{Per finger} (test, mean Dice; 11, 20, 18, 18 and 16 nails). Thumbs face sideways on a flat hand, so
they gain most over the whole-hand renders, which see the hand from above.}
\label{tab:app_finger}
\setlength{\tabcolsep}{2.8pt}
\begin{tabularx}{\linewidth}{@{}lCCCCC@{}}
\toprule
& \multicolumn{5}{c}{Mean Dice$\uparrow$} \\
\cmidrule(l){2-6}
Method & Thumb & Index & Middle & Ring & Pinky \\
\midrule
\oursrow \textbf{FrameSeg3D (Ours)} & \textbf{0.967} & \textbf{0.969} & \textbf{0.975} & \textbf{0.974} & \textbf{0.963} \\
DGCNN in the frame & \underline{0.964} & 0.962 & \underline{0.974} & 0.970 & \underline{0.960} \\
PTv3 in the frame & 0.962 & 0.939 & \textbf{0.975} & \underline{0.972} & \textbf{0.963} \\
Whole-hand render, 0.25\,mm/px & 0.957 & \underline{0.964} & 0.969 & \underline{0.972} & 0.957 \\
Whole-hand render, 0.08\,mm/px & 0.911 & 0.955 & 0.968 & 0.971 & 0.956 \\
PTv3, whole hand & 0.950 & 0.947 & 0.965 & 0.951 & 0.944 \\
ToothGroupNetwork & 0.689 & 0.796 & 0.856 & 0.864 & 0.731 \\
\bottomrule
\end{tabularx}
\end{table}

\textbf{Per finger.}
\Cref{tab:app_finger} splits the test nails by finger. FrameSeg3D is best or tied on every finger. Its gain is largest
on the thumb, which faces sideways on a flat hand: 0.967, against 0.957 for the whole-hand render at 0.25\,mm/px and
0.911 at 0.08\,mm/px.

\begin{figure*}[!htbp]
\centering
\includegraphics[width=\linewidth]{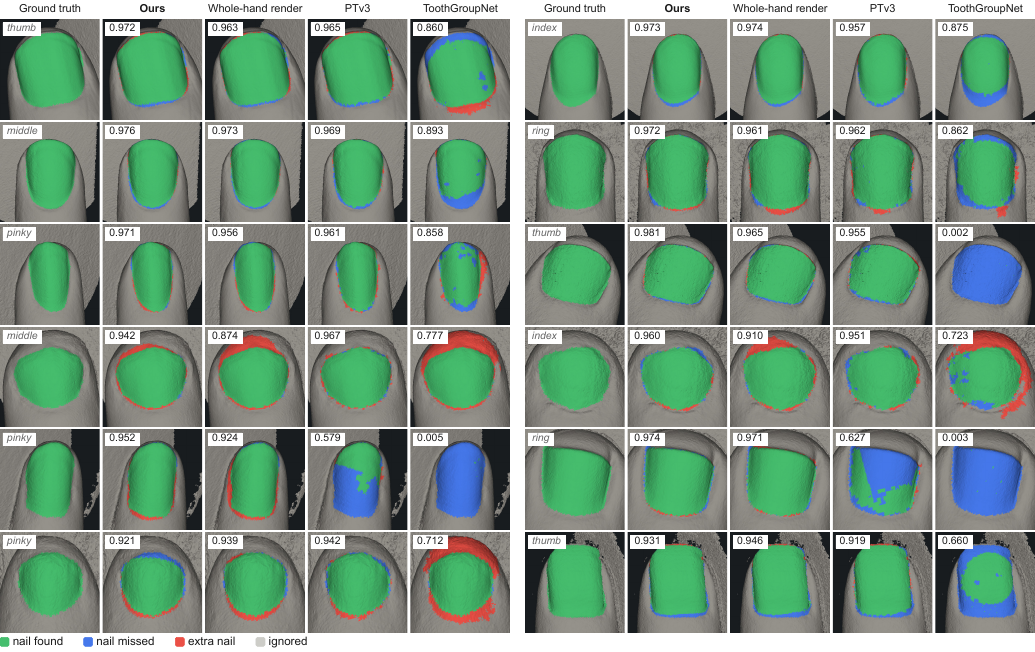}
\caption{\textbf{More test nails} with each method's Dice, picked automatically from the per-nail scores. In reading
order: a typical nail of each finger (every method near its median), the nail where ToothGroupNetwork loses most against
FrameSeg3D, the two where the whole-hand render loses most, the two where PTv3 loses most, and our two lowest nails. Each
tile is framed to its nail. Green: nail found; blue: nail missed; red: extra nail.}
\label{fig:qual_full}
\end{figure*}

\textbf{More examples.}
\Cref{fig:qual_hands_full} shows whole test hands and \cref{fig:qual_full} single nails, picked automatically from the
scores: typical cases and the cases where each baseline loses most against FrameSeg3D, so no example is chosen by hand.

\textbf{Further ablations.}
With ground-truth frames, the visibility tolerance of the lifting matters little up to 0.5\,mm (mean Dice 0.970 at 0.1,
0.25 and 0.5\,mm). At 1\,mm, BF$_{0.25}$ drops from 0.835 to 0.817, because some skin next to the nail is lifted
onto it. Training the frame network with holes and tilts (\cref{sec:app_impl}) instead of noise alone leaves the mean Dice
on clean test scans unchanged (0.970), but lowers the normal error of its frames on test hands (median $4.0^\circ$ vs.\
$5.0^\circ$, 90th percentile $9.3^\circ$ vs.\ $11.5^\circ$) and finds more teeth with the right number on the
challenge test set (97.9\% vs.\ 97.6\%).

\FloatBarrier
\section{More results on teeth}
\label{sec:app_teeth}

\textbf{The frame on teeth.}
Teeth tilt little from the jaw axis (median $14^\circ$, 95\% within $34^\circ$; thumb nails sit at $56^\circ$ from the
mat). So the tooth normal adds little. With ground-truth centers, views around the tooth normal instead of the jaw axis
add 0.15 per-tooth IoU and 0.83 boundary IoU. Most of the frame's effect on teeth comes from spending resolution on one
part. Over the whole-jaw render at 0.25\,mm/px, FrameSeg3D adds 2.59 per-tooth IoU and 14.3 boundary IoU. Of the 14.3, 7.7
come from the finer pixels and 6.6 from the frame at the same resolution. From 0.25 to 0.08\,mm/px, the whole jaw gains
0.88 IoU and 7.73 boundary IoU. The gain over the whole-jaw render grows above $30^\circ$ of tilt and is
largest on molars (\cref{tab:app_tilt}), the one tooth type where the finer whole-jaw render is worse than the coarse
one.

\begin{table}[!htbp]   %
\centering\small
\caption{\textbf{By tooth tilt and type} (test fold, per-tooth IoU per ground-truth tooth). Tilt: angle between the
tooth's normal and the jaw axis. $\Delta$: gain of FrameSeg3D in per-tooth IoU / boundary IoU at 0.25\,mm, over
ToothGroupNetwork (TGN) and over the whole-jaw render at 0.08\,mm/px.}
\label{tab:app_tilt}
\setlength{\tabcolsep}{2.4pt}
\begin{tabularx}{\linewidth}{@{}lrOCCC@{}}
\toprule
& & \multicolumn{2}{c}{Per-tooth IoU$\uparrow$} & \multicolumn{2}{c}{Gain$\uparrow$, IoU / BIoU} \\
\cmidrule(lr){3-4}\cmidrule(l){5-6}
Teeth & \multicolumn{1}{c}{$n$} &   %
  FrameSeg3D (Ours) & TGN & $\Delta$ TGN & $\Delta$ whole jaw \\
\midrule
tilt $0$--$10^\circ$ & 735 & 96.77 & 96.37 & +0.40 / +0.6 & +1.72 / +6.4 \\
tilt $10$--$20^\circ$ & 991 & 96.61 & 96.15 & +0.46 / +0.7 & +1.34 / +6.1 \\
tilt $20$--$30^\circ$ & 452 & 96.15 & 95.41 & +0.74 / +1.5 & +1.93 / +6.7 \\
tilt $>30^\circ$ & 196 & 95.54 & 94.80 & +0.73 / +1.4 & +2.97 / +9.8 \\
\midrule
incisors & 719 & 96.00 & 95.71 & +0.28 / +0.6 & +1.84 / +7.5 \\
canines & 350 & 96.37 & 96.08 & +0.29 / +0.9 & +1.15 / +5.2 \\
premolars & 705 & 96.82 & 96.53 & +0.28 / +0.5 & +1.15 / +6.0 \\
molars & 600 & 96.73 & 95.53 & +1.21 / +1.6 & +2.53 / +7.0 \\
\bottomrule
\end{tabularx}
\end{table}

\textbf{Frame prediction on teeth.}
A ground-truth tooth counts as named correctly if the frame with its number lies within 3\,mm of its centroid. On the
test fold, 99.6\% of the teeth are named correctly, with 20 extra detections in 180 scans. The median center error is
0.52\,mm (90th percentile 1.24), and the median normal error is $5.4^\circ$ (90th percentile $12.1^\circ$). On the
challenge test set, 97.9\% are named correctly, with 109 extra detections in 600 scans. There, 1.6\% of the teeth have
no frame within 3\,mm. Where the nearest frame has the wrong number, the number is off by one, that is, it names the
neighboring tooth, in 404 of 431 cases, as is also the case for ToothGroupNetwork.

\textbf{Teeth by position.}
On the challenge test set, ToothInstanceNet's lead comes from the back molars (\cref{tab:app_position}). Of its 0.39
points of per-tooth IoU, third molars account for 0.42 and second molars for 0.13, while FrameSeg3D is 0.15 ahead on teeth 1
to 6. Back molars are rarer in training. Per scan, the training fold has 0.01 third molars, the validation and test
folds 0.02, and the challenge test set 0.25. For second molars, the numbers are 1.25, 1.47, 1.32 and 1.79. Of the 87
teeth where our first run scores below 0.5 IoU, 84 have no frame, and 41 of these are third molars. The validation fold
does not show this failure: it holds 3 third molars, and every tooth in it gets a frame.

\begin{table}[!htbp]   %
\centering\small
\caption{\textbf{By tooth position on the challenge test set}: 1--6 are incisors to first molars, 7 second molars and 8
third molars. IoU: per-tooth IoU; BIoU: boundary IoU at 0.25\,mm; Fail: teeth with IoU below 0.5 (\%). Ours: mean over 3 seeds.
TGN: the released ToothGroupNetwork checkpoint.}
\label{tab:app_position}
\setlength{\tabcolsep}{2.2pt}
\begin{tabularx}{\linewidth}{@{}lrOOOCCCC@{}}
\toprule
& & \multicolumn{3}{c}{FrameSeg3D (Ours)} & \multicolumn{3}{c}{ToothInstanceNet} & TGN \\
\cmidrule(lr){3-5}\cmidrule(lr){6-8}\cmidrule(l){9-9}
Teeth & \multicolumn{1}{c}{$n$} &   %
  IoU$\uparrow$ & BIoU$\uparrow$ & Fail$\downarrow$ & IoU$\uparrow$ & BIoU$\uparrow$ & Fail$\downarrow$ & Fail$\downarrow$ \\
\midrule
1--6 & 7048 & 96.52 & 72.94 & 0.3 & 96.34 & 70.27 & 0.2 & 2.4 \\
7 & 1074 & 95.00 & 65.98 & 2.0 & 96.00 & 62.81 & 0.5 & 10.2 \\
8 & 152 & 70.00 & 46.22 & 27.6 & 92.67 & 51.79 & 1.3 & 65.8 \\
\midrule
All & 8274 & 95.84 & 71.55 & 1.1 & 96.23 & 68.96 & 0.3 & 4.6 \\
\bottomrule
\end{tabularx}
\end{table}

\begin{figure*}[!htbp]
\centering
\includegraphics[width=\linewidth]{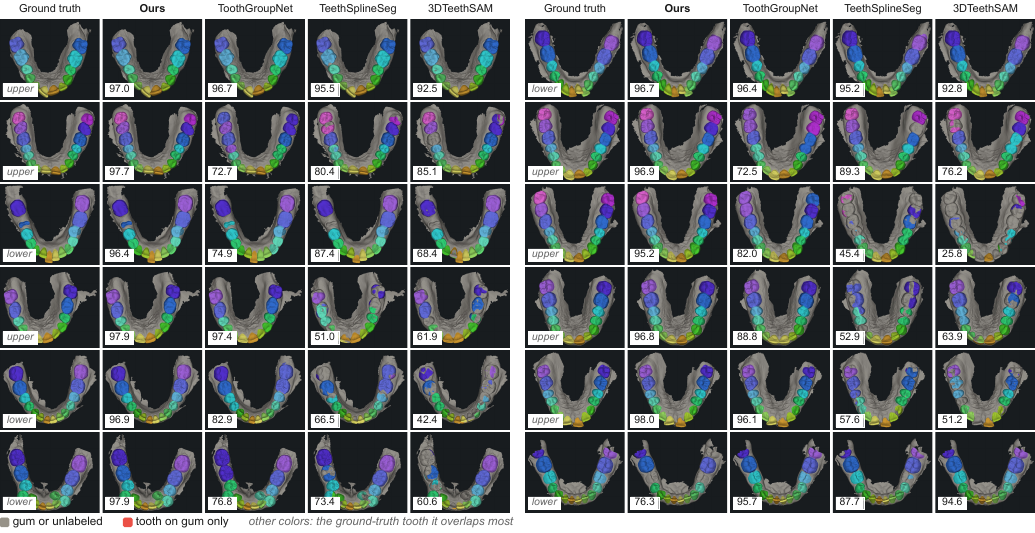}
\caption{\textbf{More jaws from the challenge test set} with each method's per-tooth IoU, picked automatically from the
per-scan scores. In reading order: a typical upper and lower jaw (every method near its median); for ToothGroupNetwork,
TeethSplineSeg and 3DTeethSAM, the three scans where each loses most against FrameSeg3D (different patients, among scans
where FrameSeg3D reaches 95); and our lowest scan, where FrameSeg3D misses three teeth (both second premolars and one
second molar). Colors as in \cref{fig:qual}.}
\label{fig:qual_teeth_full}
\end{figure*}

\textbf{Why 3DTeethSAM drops on the challenge test set.}
From the test fold to the challenge test set, 3DTeethSAM loses 6.4 per-tooth IoU, against 3.3 for ToothGroupNetwork, 2.5
for TeethSplineSeg and 0.6 for FrameSeg3D, while ToothInstanceNet gains 0.3. The drop is broad: 110 of the 600 scans fall
below 0.8 (ToothGroupNetwork 30, FrameSeg3D 3 to 6 per seed, ToothInstanceNet 0). On 94 of those 110 scans, neighboring
teeth share one predicted tooth (299 of 1{,}584 ground-truth teeth match the same predicted tooth as a neighbor). Parts
of teeth are also labeled as gum. Naming fails on the same scans (TIR 0.66). \Cref{fig:qual_teeth_full} shows the scans
where each of these methods loses most against FrameSeg3D, picked automatically from the per-scan scores.

\FloatBarrier
\section{One network and few labels}
\label{sec:app_labels}

\textbf{Few labels.}
The subsets of \cref{tab:app_labels,tab:app_labels_teeth} are nested and the same for every method: 1, 3 and 8 people hold 1, 5 and 14 of the 57 training hands, and 5
and 42 patients hold 10 and 84 of the 840 training scans. Every model selects its checkpoint on the full validation set.
We retrain both stages of ToothGroupNetwork with its official code and settings on the 5 and 42 patients. Each epoch
repeats the subset up to the 840 scans of the full fold, so a subset gets the same schedule and number of steps as the
released checkpoints (60 epochs). We keep the checkpoint with the lowest loss on the 180 validation scans.
With both stages on 3 people, the whole loss comes from two nails on two test hands, which get no frame of their own and
score close to 0. With one person, the frame network names only 83\% of the test nails correctly, and its 90th-percentile
center error is 98\,mm. With both stages on 5 patients, the median scan is unchanged, and 4 of 180 scans lose more than
5 points. On teeth, the segmentation network on 5 patients still beats ToothGroupNetwork in boundary IoU (70.18 vs.\
69.59) and ties it in their IoU (96.00 vs.\ 95.99), with 84 times fewer training scans (10 instead of 840). Retrained
on a subset, ToothGroupNetwork learns both of its stages from it, so we compare it with FrameSeg3D with both stages
trained on the same patients. On 5 patients,
ToothGroupNetwork reaches 49.80 boundary IoU and 88.34 in its own IoU, and it identifies 86.79\% of the teeth (TIR);
FrameSeg3D reaches 70.08, 95.78 and 96.66\%. On 42 patients, ToothGroupNetwork reaches 53.65, 90.64 and 94.69\%, and
FrameSeg3D 70.26, 95.88 and 97.84\%.

\textbf{Cross-domain initialization.}
Starting the segmentation network from the other domain's network instead of ImageNet weights does not help (one seed
each): nails from 3 people 0.967 vs.\ 0.966, from one person 0.939 vs.\ 0.949, and teeth from 5 patients 96.40 vs.\
96.38 per-tooth IoU (the other-domain start first in each pair).

\FloatBarrier
\end{document}